\documentclass{article}

\usepackage[main,final]{neurips_2026}

\usepackage[utf8]{inputenc} 
\usepackage[T1]{fontenc}    
\usepackage[hidelinks]{hyperref}       
\usepackage{url}            
\usepackage{booktabs}       
\usepackage{amsfonts}       
\usepackage{nicefrac}       
\usepackage{microtype}      
\usepackage[table]{xcolor}  

\usepackage{titletoc}
\usepackage{tcolorbox}
\usepackage{colortbl}
\usepackage{amsmath}
\usepackage{amssymb}
\usepackage{mathtools}
\usepackage{amsthm}
\newtheorem{theorem}{Theorem}  

\usepackage{multirow} 
\usepackage{wrapfig}
\usepackage{subcaption}
\usepackage{graphicx}
\usepackage{enumitem}
\usepackage{amsmath}
\usepackage{algorithm}
\usepackage{algorithmic}
\newcommand{\ProjectName}{MoE-SpAc}
\renewcommand{\emph}[1]{\textit{#1}}

\newcommand{\lsh}[1]{#1}

\title{\ProjectName{}: Efficient \underline{MoE} Inference Based on \underline{Sp}eculative \underline{Ac}tivation Utility in Heterogeneous Edge Scenarios}

\author{
Shuhuai Li \\
Shanghai Jiao Tong University \\
\texttt{lishuhuai@sjtu.edu.cn}
\And
Jianghao Lin\thanks{Corresponding author.} \\
Shanghai Jiao Tong University \\
\texttt{linjianghao@sjtu.edu.cn}
\AND
Dongdong Ge \\
Shanghai Jiao Tong University \\
\texttt{ddge@sjtu.edu.cn}
\And
Yinyu Ye \\
Shanghai Jiao Tong University \\
\texttt{yinyuye@sjtu.edu.cn}
}

\begin{document}

\maketitle

\begin{abstract}
Mixture-of-Experts (MoE) models enable scalable performance but face severe memory constraints on edge devices. Existing offloading strategies struggle with I/O bottlenecks due to the dynamic, low-information nature of autoregressive expert activation. 
In this paper, we propose to repurpose Speculative Decoding (SD) not merely as a compute accelerator, but as an informative \textit{lookahead sensor} for memory management, supported by our theoretical and empirical analyses. 
Hence, we introduce \textbf{\ProjectName{}}, an MoE inference framework that integrates a \lsh{training-free} \textit{Speculative Utility Estimator} to track expert demand, a \textit{Heterogeneous Workload Balancer} to dynamically partition computation via online integer optimization, and an \textit{Asynchronous Execution Engine} to unify the prefetching and eviction in the same utility space.
Extensive experiments on seven benchmarks demonstrate that \ProjectName{} achieves a \textbf{42\%} improvement in TPS over the SOTA SD-based baseline, and an average \textbf{4.04$\times$} speedup over all standard baselines.
Code is available at \url{https://github.com/lshAlgorithm/MoE-SpAc}.
\end{abstract}

\section{Introduction}

Large Language Models (LLMs) have achieved remarkable capabilities by scaling to hundreds of billions or even trillions of parameters~\citep{brown2020language, ouyang2022training, achiam2023gpt}. The Mixture-of-Experts (MoE) architecture has been instrumental in this growth, offering a path to vastly larger models while keeping the computational cost manageable~\citep{jacobs1991adaptive, zhou2022mixture, roller2021hash}. By routing each input token through only a small subset of expert networks, MoE dramatically reduces the required floating-point operations (FLOPs), compared to dense models of similar size.


However, this parameter efficiency imposes a severe memory penalty. The immense parameter footprint creates a critical barrier to deployment in resource-constrained environments, such as personal devices and edge hardware, besides 
To this end, \textit{heterogeneous offloading} has become the standard solution~\citep{pan2026trimoe, zhu2026dali}.
Typically, the full set of expert weights resides in high-capacity CPU memory, while activated experts are transferred on-demand to the high-throughput GPU device.

\begin{figure}[t]
\centering
\includegraphics[width=1.00\textwidth]{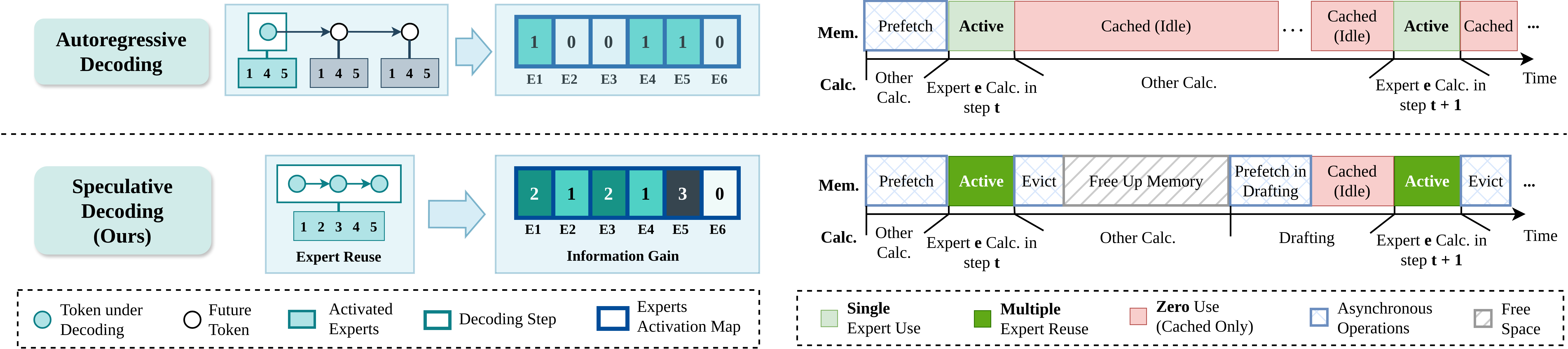}
\caption{
Two key observations how speculative decoding (SD, Bottom) better benefits MoE inference than autoregressive decoding (AR, Top).
\textbf{Theoretically} (Left), SD enables expert reuse and transforms activation utility signals from AR into informative frequency-valued ones. 
\textbf{Practically} (Right), experts can be scheduled in parallel by proactively leaving more memory for experts to come and asynchronously prefetching experts during the drafting phase.
\vspace{-1.5em}
}
\label{fig:ob}
\end{figure}

Existing offloading strategies generally fall into two categories.
\emph{The first} strategy involves \textbf{GPU intensive} expert calculation. It offloads activated expert weights from CPU to GPU on demand, which significantly exacerbates the I/O bottleneck between heterogeneous devices~\citep{fang2025accurate}. 
While predictive \textit{prefetching} attempts to hide this latency using auxiliary networks~\citep{du2024sida, hwang2024pre} or historical activation\footnote{In this paper, unless otherwise specified, ``activation'' refers to the MoE gating decision (i.e., which experts are selected).} patterns~\citep{li2023accelerating, xue2024moe, madan2026speculating}, these methods suffer from unavoidable prediction errors due to the binary, low-information signals during the autoregressive (AR) generation.
\emph{The second} strategy is \textbf{CPU-GPU hybrid} expert calculation, which allows CPU execution of experts that are missed in GPU VRAM to reduce expert loading overhead. Fiddler~\citep{kamahori2024fiddler} and kTransformers~\citep{10.1145/3731569.3764843} adopt this method but rely on static expert allocation or profiling, failing to capture the dynamic nature of expert activation~\citep{yu2025prescope}. 
Although HybriMoE~\citep{zhong2025hybrimoe} introduces a dynamic strategy, its prefetching and caching algorithms remain decoupled, preventing a unified scheduling objective and resulting in suboptimal load balancing~\citep{chen2026universal}, more relevant works are attached in Appendix~\ref{apix:related_work}.

In this paper, we identify a novel synergy between speculative decoding (SD)~\citep{leviathan2023fast, xia2022speculative} and heterogeneous MoE inference. 
We propose to repurpose SD not merely as an accelerator, but as an informative lookahead sensor for memory management.
Typically, SD accelerates inference by using a smaller draft model to generate candidate tokens, which the larger target model verifies in a parallel forward pass. 
On top of that, we observe that the SD paradigm offers decisive advantages for MoE inference from both theoretical and practical aspects.

\textbf{Theoretically}, SD \emph{reuses} the expert weights and \emph{enriches} the transmission of expert activation signals. 
    As shown in Figure~\ref{fig:ob}(Left), while autoregressive decoding yields a binary indicator (activated or not) for a single step. 
    Hence, SD transforms the low-information binary signals of autoregressive decoding into non-binary, informative signals for expert scheduling. 
    Furthermore, SD introduces inherent \emph{fault tolerance}: exact frequency prediction is unnecessary; coarse-grained utility estimation is sufficient to guide effective scheduling policies.

\textbf{Practically}, as shown in Figure~\ref{fig:ob}(Right), SD enables \emph{parallel scheduling}, where experts can be prefetched in I/O while the GPU generates draft tokens. 
Additionally, SD presents hot and cold experts in a single step to optimize \emph{heterogeneous compute utilization}: 
by offloading low-frequency experts to the CPU (sequential processing), we reserve the high-throughput GPU for high-frequency experts like E5 on left of Figure~\ref{fig:ob} (parallel processing), thereby balancing the load effectively across heterogeneous devices.

Leveraging these insights, we propose \textbf{\ProjectName{}}, an efficient \textbf{MoE} inference framework based on \textbf{sp}eculative \textbf{ac}tivation utility in heterogeneous edge scenarios.
Rather than treating SD merely as a computation accelerator, \ProjectName{} introduces a training-free \textit{Speculative Utility Estimator}. This estimator evaluates expert demand in a stable, discrete utility space via an inertial transition mechanism.
Guided by these utilities, a \textit{Heterogeneous Workload Balancer} solves an online integer optimization problem at each layer, dynamically determining a global threshold to partition experts between GPU and CPU based on real-time I/O and memory constraints.
These decisions are executed by an \textit{Asynchronous Execution Engine}, which unifies prefetching and eviction under the same utility metric to minimize synchronization overhead.
In this way, \ProjectName{} not only masks I/O latency but also harmonizes the computational load across heterogeneous hardware, transforming the memory wall into a manageable expert scheduling problem.

In summary, our contributions are as follows:

\begin{itemize}[leftmargin=10pt]
    \item \textbf{Paradigm Shift:} 
    We redefine the role of speculative decoding in MoE inference, shifting its paradigm from a mere computation accelerator to an informative \textit{lookahead sensor} for memory management, which is supported by both theoretical and empirical analysis. 

    \item \textbf{Unified Scheduling Framework.} 
    We propose \textit{MoE-SpAc}, which integrates speculative decoding for online heterogeneous expert scheduling based on unified expert utilities. 
    \ProjectName{} dynamically harmonizes CPU-GPU workloads, ensuring optimal throughput by adapting to strict I/O and memory constraints in real-time.

    \item \textbf{SOTA Performance:} 
    Extensive experiments on seven benchmarks demonstrate that \ProjectName{} achieves a \textbf{42\%} speedup in TPS over the best SD-based baseline, and an average \textbf{4.04$\times$} speedup over all standard baselines. 
\end{itemize}

\section{Preliminary and Problem Formulation} \label{sec:preliminary}

\subsection{Speculative Decoding} \label{sec:prelim_sd}

Speculative Decoding (SD) accelerates inference by leveraging a lightweight \textit{draft model} to generate a sequence of $\gamma$ candidate tokens, which are verified in parallel by the larger \textit{target model}. 
This parallelism maximizes GPU compute utilization compared to the serial nature of autoregressive decoding, generating $[1, \gamma + 1]$ tokens at once.
Let $\alpha$ denote the acceptance probability of a draft token. The expected number of tokens generated per decoding step, $\Omega(\gamma, \alpha)$, is given by \citet{leviathan2023fast}:
\begin{equation}\label{for:omega}
\Omega(\gamma,\alpha) := \mathbb{E}[\text{\# Generated Tokens}] = \frac{1-\alpha^{\gamma+1}}{1-\alpha}.
\end{equation}

The latency of generating $\Omega(\gamma,\alpha)$ tokens using speculative decoding ($T_{SD}$) or autoregressive decoding ($T_{AR}$) is:
\begin{equation}
\begin{aligned}
T_{SD} &= \gamma \cdot T_D + T_V, \\
T_{AR} &= \Omega(\gamma, \alpha) \cdot T_V,
\end{aligned}
\end{equation}
where $T_D$ is the time elapsed for the draft model to generate a single token, and $T_V$ is the latency of a single parallel verification pass by the target model. 
Hence, the wall-clock speedup factor is $T_{AR}/T_{SD} = \frac{1 - \alpha^{\gamma + 1}}{(1 - \alpha)(\gamma c + 1)}$, where $c = T_D / T_V$ represents the cost coefficient.

\subsection{Mixture-of-Experts Architecture}

Unlike dense models that utilize all parameters for every forward pass, MoE models conditionally activate only a subset of parameters per token.
Formally, let $\mathbf{h} \in \mathbb{R}^d$ denote the input hidden state for a token at a specific MoE layer. This layer contains a set of $N$ experts, $\mathcal{E} = \{E_1, E_2, \dots, E_N\}$, where each expert $E_i(\cdot)$ is typically a feed-forward network (FFN).
The output $\mathbf{y}$ is the weighted aggregation of activated experts under a Top-$k$ ($k \ll N$) routing strategy:
\begin{equation}
\begin{aligned}
    \mathbf{s}&=\mathbf{W}_g \mathbf{h} \in \mathbb{R}^{N}\\
    \mathcal{T}(\mathbf{h}) &= \text{Top-}k(\mathbf{s}) \subset\{1,\dots,N\},\\
    g_i(\mathbf{h})&=\frac{\exp(\mathbf{s}_i)}{\sum_{j\in \mathcal{T}(\mathbf{h})}\exp(\mathbf{s}_j)}, \;\; i\in \mathcal{T}(\mathbf{h}),\\
    \mathbf{y} &= \sum\nolimits_{i \in \mathcal{T}(\mathbf{h})} g_i(\mathbf{h}) \cdot E_i(\mathbf{h}), \label{eq:moe_output}
\end{aligned}
\end{equation}
where $\mathbf{W}_g \in \mathbb{R}^{N \times d}$ is the learnable routing matrix, $\mathcal{T}(\mathbf{h})$ is the indices of $k$ activated experts. and $g_i(\mathbf{h})$ is the normalized weight for expert $E_i$.
Thus, the majority ($N - k$) of experts remain inactive, creating significant sparsity.

\subsection{Online Heterogeneous Expert Scheduling}
\label{sec:task formulation}

We formalize the efficient inference of MoE models on memory-constrained edge devices as an online decision-making problem, termed \textit{Heterogeneous Expert Scheduling}. 
The objective is to dynamically prioritize high-frequency experts (hot) to GPU while offloading the computation of low-frequency experts (cold) to CPU at each inference step to minimize the I/O-induced latency.

\textbf{Inference Step.} 
Let $t$ denote a discrete inference step to generate a sequence of tokens $\mathbf{X}_t$. 
In autoregressive (AR) decoding, a step $t$ corresponds to the generation of a single token, i.e., $|\mathbf{X}_t| = 1$. 
In speculative decoding (SD), a step $t$ corresponds to a verification cycle that potentially generates multiple tokens, i.e., $\mathbb{E}|\mathbf{X}_t| = \Omega(\gamma,\alpha) \ge 1$.

\textbf{Activation Frequency.} 
For a given step $t$ and input tokens $\mathbf{X}_t$, we define the \textit{observed} activation frequency $f_{i,t}$ for expert $E_i$ as the cumulative number of times that $E_i$ is activated in Eq.~\ref{eq:moe_output} over tokens $\mathbf{X}_t$.
For AR, $f_{i,t} \in \{0, 1\}$, providing a low-information, binary signal. 
For SD, $f_{i,t} \in [0, \gamma+1]$, providing a more informative, non-binary signal that reflects the \textit{intensity} of expert demand.

\textbf{Activation Utility Estimation.}
To guide the expert scheduling, it requires a metric of expert priority for the \textit{upcoming} step $t+1$. 
Let \lsh{$u_{i, t+1}$} denote the \textit{ground-truth utility}, which is determined by a monotonically non-decreasing mapping function $G(\cdot)$ applied to the future (unknown) frequency:
\begin{equation}
    u_{i, t+1} = G(f_{i, t+1}),
\end{equation}
where $G(\cdot)$ maps frequencies to a discrete utility space (e.g., priority levels). 
Since $f_{i, t+1}$ is unobservable at step $t$, we must estimate the future utility $\hat{u}_{i, t+1}$ using a scoring function $F$ based on the current utility and historical frequencies:
\begin{equation}\label{eq:score_update}
    \hat{u}_{i, t+1} = F\left(\hat{u}_{i, t}, \{f_{i, j}\}_{j=1}^{t}\right).
\end{equation}

\textbf{Heterogeneous Resource Allocation.}
The final scheduling decision is a thresholding operation on estimated utilities. 
Let $\tau_t$ be a dynamic threshold determined by the system's resource constraints at step $t$. 
We partition the experts as:
\begin{itemize}[leftmargin=10pt]
    \item \textbf{Hot Experts} ($\hat{u}_{i, t+1} \ge \tau_t$): Experts with high activation frequency across the speculative window should be prioritized (prefetched) for high-throughput GPU.
    \item \textbf{Cold Experts} ($\hat{u}_{i, t+1} < \tau_t$): Experts with negligible or zero activation should be evicted from GPU to CPU-side execution for load balancing.
\end{itemize}

\lsh{To make the scheduling decision with more reliable information, we theoretically demonstrate that beyond simple speedup,} SD fundamentally alters the nature of the online heterogeneous expert scheduling by providing \textit{Expert Reuse}, \textit{Information Gain}, and \textit{Fault Tolerance} in Appendix~\ref{app:theretical analysis on SD}.

\begin{figure*}[t] 
\begin{center}
\includegraphics[width=0.995\textwidth]{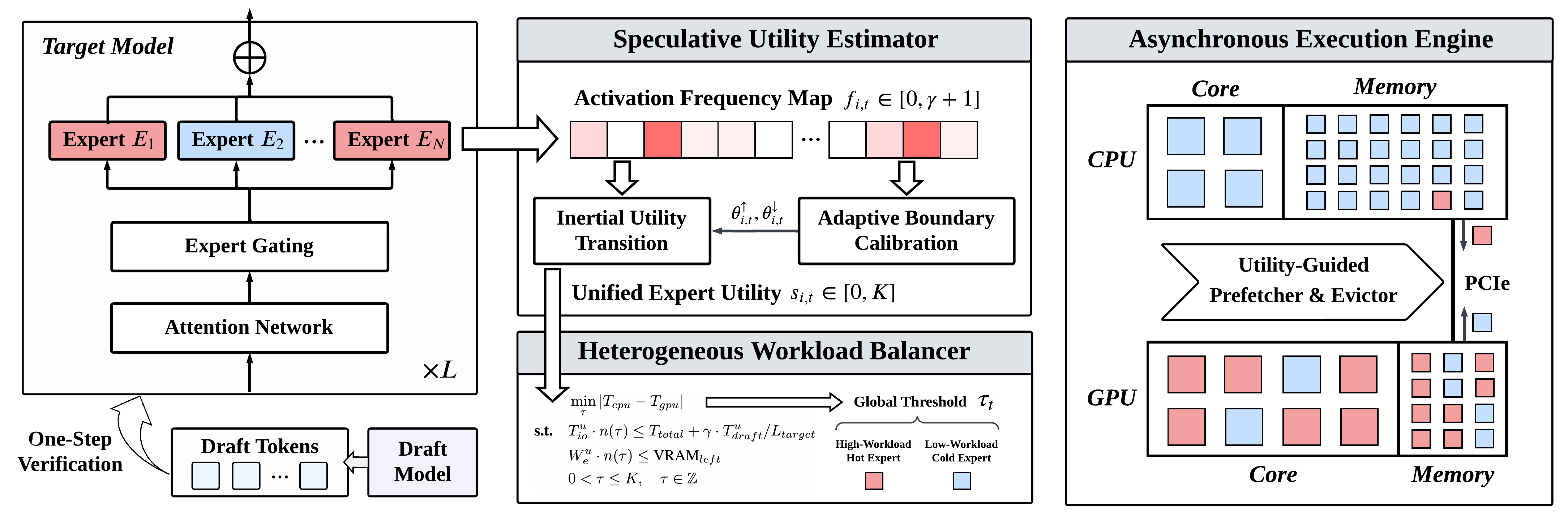}
\end{center}
\caption{
Overall framework of \ProjectName{}.
} 
\label{fig:framework}
\end{figure*}
\section{\ProjectName{}}\label{sec:\ProjectName{}}
\subsection{Overview}\label{sec:overview}

To resolve the online heterogeneous expert scheduling problem, we present \textbf{\ProjectName{}}, an efficient \textbf{MoE} inference framework based on \textbf{Sp}eculative \textbf{Ac}tivation utility for edge scenarios. 

As shown in Figure~\ref{fig:framework}, \ProjectName{} consists of three key components: 
\textbf{(1)} \textit{Speculative Utility Estimator} predicts future expert demand for each verification step based on  inertial utility transition and adaptive boundary calibration; 
\textbf{(2)} \textit{Heterogeneous Workload Balancer} harmonizes computation loads between devices by solving layer-wise online integer optimization to determine the optimal global threshold for GPU-CPU expert partitioning; 
and \textbf{(3)} \textit{Asynchronous Execution Engine} actualizes prefetching and eviction decisions based on a \textit{unified} utility metric, managing I/O operations asynchronously without stalling the critical computation.
In this way, we adapt speculative decoding to MoE inference in edge scenarios, not merely as a compute accelerator, but also as a core lookahead sensor for critical memory management between heterogeneous devices.

\subsection{Speculative Utility Estimator}\label{sec:predictor}
\begin{wrapfigure}{r}{0.5\textwidth}
    \begin{minipage}{0.5\textwidth}
    \vspace{-2em}  
    \begin{algorithm}[H]
      \caption{Speculative Utility Estimation}
      \label{algo:merged_sm}
      \begin{algorithmic}[1]
        \STATE {\bfseries Hyperparameters:} Utility Upper Bound $K$, Number of Experts $N$, Forgetting Factor $\lambda$.
        \STATE {\bfseries Initialize:} Utility $u_{i,1} = 0$ for all $i \in \{1, \dots, N\}$.
        \STATE {\bfseries Initialize:} Boundaries $\theta^{\uparrow}_{i,1} = \theta^{\downarrow}_{i,1} = \lfloor \gamma/2 \rfloor$.
        
        \FOR{inference step $t = 1, 2, \dots$}
          \STATE Observe activation frequencies $\mathbf{f}_t = \{f_{1,t}, \dots, f_{N,t}\}$
          
          \FOR{each expert $i \in \{1, \dots, N\}$}
            \STATE $\Delta_{i,t} \leftarrow f_{i,t} - f_{i,t-1}$
            
            \STATE \textcolor{teal}{// Inertial Utility Transition}
            \IF{$\Delta_{i,t} \geq \theta^{\uparrow}_{i,t}$}
                \STATE $u_{i,t+1} \leftarrow \min(K, u_{i,t} + 1)$
            \ELSIF{$-\Delta_{i,t} \geq \theta^{\downarrow}_{i,t}$}
                \STATE $u_{i,t+1} \leftarrow \max(0, u_{i,t} - 1)$
            \ELSE
                \STATE $u_{i,t+1} \leftarrow u_{i,t}$
            \ENDIF
            
            \STATE \textcolor{teal}{// Adaptive Boundary Calibration}
            \IF{$\Delta_{i,t} > 0$}
                \STATE $\theta^{\uparrow}_{i,t+1} \leftarrow \lfloor (1 - \lambda)\theta^{\uparrow}_{i,t} + \lambda \cdot \Delta_{i,t} \rfloor$
                \STATE $\theta^{\downarrow}_{i,t+1} \leftarrow \theta^{\downarrow}_{i,t}$
            \ELSIF{$\Delta_{i,t} < 0$}
                \STATE $\theta^{\uparrow}_{i,t+1} \leftarrow \theta^{\uparrow}_{i,t}$
                \STATE $\theta^{\downarrow}_{i,t+1} \leftarrow \lfloor (1 - \lambda)\theta^{\downarrow}_{i,t} + \lambda \cdot |\Delta_{i,t}| \rfloor$
            \ELSE
                \STATE $\theta^{\uparrow}_{i,t+1} \leftarrow \theta^{\uparrow}_{i,t}; \quad \theta^{\downarrow}_{i,t+1} \leftarrow \theta^{\downarrow}_{i,t}$
            \ENDIF
          \ENDFOR
        \ENDFOR
      \end{algorithmic}
    \end{algorithm}
    \end{minipage}
    \vspace{-2em}
\end{wrapfigure}

To effectively guide the heterogeneous scheduling, we have to derive an activation utility that is both reflective of \textit{next-step demand} and robust against \textit{frequency fluctuations}. 
Due to the temporal locality and smooth evolving of the accumulated activation frequencies across speculative windows, we estimate the activation utility in a compressed discrete space, i.e., $u_{i,t} \in \{0, \dots, K\}$, where $K$ is a fixed utility upper bound. We use $K=4$, following the logarithmic guideline $\lceil\log_2(f_{\max}+1)\rceil$ for the maximum per-expert window count $f_{\max}$ (eight draft tokens give $\lceil\log_2 9\rceil=4$). 
Here, $u_{i,t}=0$ indicates a dormant (extremely cold) expert, while $u_{i,t}=K$ represents a fully saturated (extremely hot) expert. 
As illustrated in Algorithm~\ref{algo:merged_sm}, we conduct \textit{speculative utility estimation} based on the historical activation frequencies at each verification step $t$, which consists of two key components: \textit{inertial utility transition} and \textit{adaptive boundary calibration}.

\textbf{Inertial Utility Transition.} 
To better estimate the activation utility dynamically, we introduce an inertial update mechanism. At each step $t$, the utility $u_{i,t}$ of expert $E_i$ remains unchanged unless the frequency fluctuation, defined as $\Delta_{i,t} = f_{i, t} - f_{i, t-1}$, exceeds a significant margin.\footnote{Since the inertial utility transition does not require fitting to a ground-truth utility, we denote the estimated utility as $u_{i,t}$, instead of $\hat{u}_{i,t}$ for simplicity.} Formally, the transition is governed by:
\begin{equation}\label{eq:utility_transition}
    u_{i, t+1} \leftarrow 
    \begin{cases} 
    \min(K, u_{i, t} + 1) & \text{if } \Delta_{i,t} \geq \theta^{\uparrow}_{i,t} \\
    \max(0, u_{i, t} - 1) & \text{if } -\Delta_{i,t} \geq \theta^{\downarrow}_{i,t} \\
    u_{i, t} & \text{otherwise}
    \end{cases}
\end{equation}
where $\theta^{\uparrow}_{i,t}$ and $\theta^{\downarrow}_{i,t}$ are expert-specific fluctuation boundaries at step $t$ to trigger the utility transition upward ($+1$) or downward ($-1$). 
This inertia ensures that the system only incurs the high I/O cost of prefetching or eviction when there is a sustained shift in expert demand, effectively filtering out high-frequency noise.

\textbf{Adaptive Boundary Calibration.}
The boundaries are initialized as $\lfloor\gamma/2\rfloor$, but should be adaptively calibrated at each step to reflect the evolving trend of expert demand inertia.
As shown in Algorithm~\ref{algo:merged_sm}, according to the positivity of frequency fluctuation, we update the boundaries via a moving average with a forgetting factor $\lambda$. 
In this way, the system adapts to the current workload characteristics, maintaining high responsiveness during distribution shifts while preserving stability during steady processes.

We provide empirical and theoretical analysis to support the design of speculative utility estimation in Appendix~\ref{app:design rationale utility estimation}.

\subsection{Heterogeneous Workload Balancer}\label{sec:balancer}

With the expert utilities estimated above, now we have to determine the optimal global threshold $\tau_t$ that separates hot experts (prioritized on GPU) from cold experts (processed on CPU). 
To this end, we model this as an online integer optimization problem, solved for \textit{each Transformer layer} of the target model at \textit{each verification step}. 
For brevity, we omit the subscripts for step $t$ and layer $l$ in the following formulation.

\begin{figure}[t]
\centering
\begin{subfigure}[t]{0.5\textwidth}
    \centering
    \scriptsize
    \begin{tabular}{l p{0.5\linewidth}}
        \toprule
            \textbf{Symbol} & \textbf{Description} \\
            \midrule
            $\tau$ & Threshold, the decision variable\\
            \midrule
            $n(\tau)$ & Number of prefetched experts \\
            \midrule
            $r_{c}(\tau)$, $r_{g}(\tau)$ & Ratio processed on CPU (sequential) and GPU (parallel), respectively. \\
            \midrule
            $T^{u}_{cpu}$,$T^{u}_{gpu}$ & Unit expert inference time on CPU, GPU \\
            \midrule
            $T^{u}_{io}$ & Unit expert I/O transmission time \\
            \midrule
            $T^{u}_{draft}$ & Unit time for drafting one token \\
            \midrule
            $L_{target}$ & Number of target model's layer \\
        \bottomrule
    \end{tabular}
\end{subfigure}
\begin{subfigure}[t]{0.47\textwidth}
    \centering
    \includegraphics[width=\linewidth]{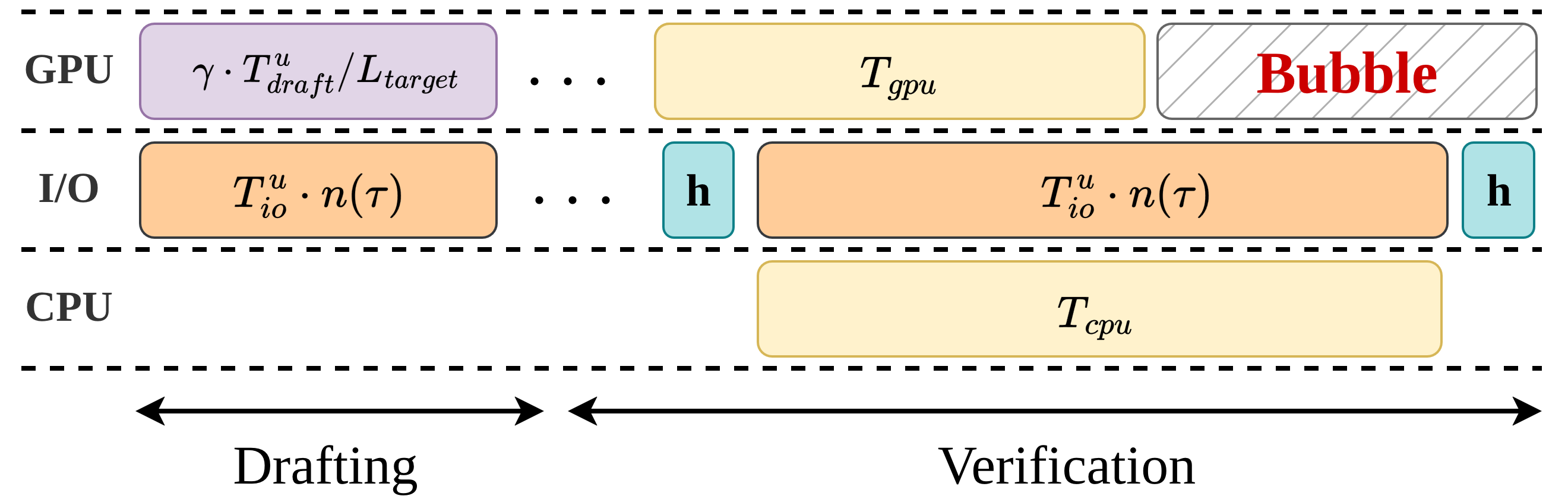}
    \captionsetup{labelformat=empty}
    \caption{Pipeline of single layer forward in SD scenario. GPU and CPU stands for calculation on each device. $h$ in I/O stands for the transmission of hidden states of tokens.}
\end{subfigure}

\caption{Notation and \ProjectName{} pipeline illustration.}\label{fig:pip-simple}
\end{figure}

\textbf{Objective.}
With notations and illustration in Figure~\ref{fig:pip-simple},
our goal is to balance the computation load between heterogeneous devices to minimize synchronization overhead (bubbles).
We can derive the total expert execution times for CPU ($T_{cpu}$) and GPU ($T_{gpu}$), respectively:
\begin{align}
    T_{cpu} &= r_c(\tau) \cdot \gamma \cdot k \cdot T^{u}_{cpu}, \label{eq:cpu_time} \\
    T_{gpu} &= r_g(\tau) \cdot b \cdot T^{u}_{gpu}. \label{eq:gpu_time}
\end{align}
The system latency is dominated by the bottleneck device: $T_{total} = \max(T_{cpu}, T_{gpu})$. 
Hence, the objective is to minimize the difference between execution times of heterogeneous devices, i.e., $|T_{cpu} - T_{gpu}|$.

\textbf{Constraints.}
The solution must satisfy the following constraints. (1) \textit{I/O Constraint}: the time used to prefetch $n(\tau)$ new experts to GPU must not exceed the available computation window.
As a bonus brought by SD, we can prefetch during the drafting phase.
(2) \textit{Memory Constraint}: the total size of newly prefetched experts must fit within the remaining VRAM.
(3) \textit{Decision Variable Constraint}: the decision variable $\tau$ should fall in the discrete utility space.

\textbf{Formulation.}
The utility bound $K$ controls the resolution of the scheduling decision space. The logarithmic guideline in Section~\ref{sec:predictor} yields $K=4$ for our speculative window. Figure~\ref{fig:in-depth analysis} (Right) shows that performance stabilizes for $K\geq4$: smaller bounds merge useful hot/cold distinctions, whereas larger bounds provide little additional benefit. We therefore use $K=4$ throughout the default configuration.

Combining the objective and constraints, the complete optimization problem is:
\begin{align}
    & \min_{\tau} |T_{cpu} - T_{gpu}| \label{eq:objective} \\
    \textbf{s.t.} \quad 
    & T^{u}_{io} \cdot n(\tau) \leq T_{total} + \gamma \cdot T^{u}_{draft} / L_{target} \label{eq:io_const} \\
    & W^{u}_e \cdot n(\tau) \leq \text{VRAM Left} \label{eq:vram_const} \\
    & 0 < \tau \leq K, \quad \tau \in \mathbb{Z} \label{eq:integer_const}
\end{align}

where $W_e^u$ stands for the memory space to store an unit expert. For an affine device-time model and a contiguous feasible interval, the optimal threshold is obtained by checking the integers adjacent to the analytical crossing and the feasible boundaries. Monotonicity alone does not imply convexity; Appendix~\ref{pip} states the conditions and fallback search. The detailed formulation and solution derivation is provided in Appendix~\ref{pip} due to the page limit.

\subsection{Unified Asynchronous Execution Engine}\label{sec:executor}

To actualize the optimal threshold $\tau$ derived by the heterogeneous workload balancer, it requires an efficient mechanism to physically migrate expert weights without stalling the computation pipeline. 
Hence, we design an \textit{asynchronous execution engine} that orchestrates prefetching and eviction based on the \textit{unified} activation utility $u_{i,t}$.

\emph{Utility-Guided Prefetching} is orchestrated via a multi-level priority queue that processes expert requests in descending order of utility (from $K$ down to the dynamic threshold $\tau_t$), while maintaining layer-wise temporal order through FIFO sub-queues. Complementarily, \emph{Utility-Guided Eviction} is handled by a state-ordered pool, which indexes all GPU-resident experts by their activation utility. This unified architecture allows the system to achieve scheduling consistency and perform high-speed cache pruning in $O(log N)$ time, ensuring that only the most critical experts occupy the limited GPU VRAM.
More implementation details are provided in Appendix~\ref{apix:impl}.

\textbf{Runtime overhead.} On Qwen3-30B-A3B, estimator state occupies approximately 30 KB and updating all $N$ experts costs $O(N)$, measured at 0.063 ms per step. Threshold selection takes 0.529 ms per 512-token generation. These operations overlap hidden-state transfers and GPU attention, respectively; state size remains fixed with generation length. Appendix~\ref{pip} details the solver assumptions.

\section{Experiments}\label{sec:exp}

\subsection{Experiment Setups}\label{sec:setup}
\label{sec:exp_setup}

\textbf{Hardware.} We evaluate \ProjectName{} under constrained resources with a single NVIDIA GeForce RTX 4090 GPU, connected to CPU memory via a PCIe 4.0 interface (32GB/s). 

\textbf{Benchmarks.}
We adopt diverse benchmarks to cover different capabilities of LLMs, including
MMLU-Pro~\citep{wang2024mmlu}, 
MT-bench~\citep{zheng2023judging}, 
HumanEval~\citep{chen2021evaluating}, 
GSM8K~\citep{cobbe2021training}, 
Alpaca~\citep{alpaca},  CNN/DailyMail~\citep{nallapati2016abstractive}, and 
QA~\citep{kwiatkowski2019natural}.

\textbf{Baselines.}
The baselines fall into two categories. 
(1) General inference engines, including \texttt{Accelerate}~\citep{wolf2019huggingface}, \texttt{vLLM}~\citep{kwon2023efficient}, and \texttt{llama.cpp}~\citep{ollama2024}. 
(2) Inference systems tailored for MoE models, including \texttt{Mixtral Offloading}~\citep{eliseev2023fast}, \texttt{MoE-Infinity}~\citep{xue2024moe}, \texttt{SP-MoE}~\citep{chen2025sp}, \texttt{Fate}~\citep{fang2025fate}, \texttt{HybriMoE}~\citep{zhong2025hybrimoe}, \texttt{MoE-SpeQ}~\citep{wang2025moe}, and \texttt{SpecExp}~\citep{madan2026speculating} (Speculating Experts).
Detailed descriptions of each baseline are given in Appendix~\ref{app:baselines}.

\textbf{Configuration Details.}
We employ the MoE model \texttt{Qwen3-30B-A3B} as the target verifier and a quantized dense model \texttt{Qwen3-4B-FP8} as the draft model~\citep{yang2025qwen3}.
We set the hyperparameters as follows: draft length $\gamma=8$, forgetting factor $\lambda=0.1$. 
The maximum generation length per request is 512 tokens. We use \texttt{--cpu-offload-gb 60} for the vLLM baseline.
Following previous 
works~\citep{xue2024moe, chen2025sp, svirschevski2024specexec}, we use a batch size of 1 to simulate edge deployment.
\lsh{Experiments on additional models and settings are presented in Section~\ref{sec:generalization}.}

\textbf{Metric.}
We adopt tokens per second (TPS) and latency (Lat.) as evaluation metrics, which capture the end-to-end efficiency of the inference system. \lsh{Exact speculative sampling preserves the target distribution when acceptance/rejection is implemented correctly. Draft quantization affects acceptance, while heterogeneous numerical consistency requires separate validation (Appendix~\ref{app:revision-evidence}).}

\subsection{Overall Performance}\label{sec:res}

\begin{table*}[t]
\centering
\caption{
The tokens per second (TPS) and latency (Lat.) performance of different inference methods. The best results are given in \textbf{bold} while the second-best values are \underline{underlined}.
\textit{Rel.Imprv.} indicates the relative improvement of \ProjectName{} against the best baseline. Dashes denote unreported measurements; averages cover all seven benchmarks.
}
\label{tab:main_result}
\label{tab:additional-baselines}
\resizebox{\textwidth}{!}{
\renewcommand\arraystretch{1.07}
\begin{tabular}{l cc cc cc cc cc cc cc cc}
\toprule
\multirow{2}{*}{\textbf{Method}} 
& \multicolumn{2}{c}{\textbf{MMLU-Pro}} 
& \multicolumn{2}{c}{\textbf{MT-bench}} 
& \multicolumn{2}{c}{\textbf{GSM8K}} 
& \multicolumn{2}{c}{\textbf{HumanEval}} 
& \multicolumn{2}{c}{\textbf{Alpaca}} 
& \multicolumn{2}{c}{\textbf{CNN/DM}} 
& \multicolumn{2}{c}{\textbf{QA}} 
& \multicolumn{2}{c}{\textbf{Average}} \\
\cmidrule(lr){2-3} \cmidrule(lr){4-5} \cmidrule(lr){6-7} \cmidrule(lr){8-9} \cmidrule(lr){10-11} \cmidrule(lr){12-13} \cmidrule(lr){14-15} \cmidrule(lr){16-17}
 & TPS$\uparrow$ & Lat.$\downarrow$ & TPS$\uparrow$ & Lat.$\downarrow$ & TPS$\uparrow$ & Lat.$\downarrow$ & TPS$\uparrow$ & Lat.$\downarrow$ & TPS$\uparrow$ & Lat.$\downarrow$ & TPS$\uparrow$ & Lat.$\downarrow$ & TPS$\uparrow$ & Lat.$\downarrow$ & TPS$\uparrow$ & Lat.$\downarrow$ \\
\midrule
Accelerate      & 1.13 & 495.37 & 1.15 & 435.55 & 1.21 & 237.45 & 1.15 & 393.60 & 1.09 & 350.83 & 1.27 & 222.47 & 1.16 & 138.42 & 1.17 & 324.81 \\
vLLM            & 2.65 & 203.74 & 2.67 & 187.88 & 2.67 & 94.63 & 2.63 & 158.46 & 2.67 & 137.72 & 2.57 & 104.28 & 2.68 & 57.43 & 2.65 & 134.88 \\
llama.cpp       & 15.17 & 37.92 & 14.34 & 36.08 & 15.01 & 19.21 & 14.74 & 30.57 & 16.86 & 22.40 & 15.42 & 19.46 & 15.09 & 11.07 & 15.23 & 25.24 \\
llama.cpp-w/ SD & \underline{17.29} & \underline{32.88} & \underline{17.58} & \underline{29.03} & \underline{19.03} & \underline{15.09} & \underline{18.78} & \underline{24.01} & \underline{18.02} & \underline{21.51} & \underline{16.46} & \underline{18.36} & \underline{16.49} & \underline{10.43} & \underline{17.66} & \underline{21.62} \\
\midrule
Mixtral Offload & 2.43 & 230.87 & 2.36 & 212.70 & 2.35 & 120.25 & 2.48 & 177.87 & 2.33 & 162.32 & 2.23 & 131.72 & 2.33 & 70.86 & 2.36 & 158.08 \\
MoE-Infinity    & 3.64 & 151.43 & 3.64 & 136.84 & 3.63 & 75.42 & 3.63 & 119.53 & 3.48 & 107.82 & 3.56 & 80.56 & 3.44 & 46.13 & 3.57 & 102.53 \\
SpecExp         & 2.96 & 172.97 & 2.94 & 174.15 & 2.94 & 174.15 & 2.93 & 174.74 & 2.87 & 178.40 & 2.94 & 174.15 & 2.83 & 180.92 & 2.92 & 175.64 \\
MoE-SpeQ        & 4.77 & 107.34 & 4.70 & 108.94 & 4.83 & 106.00 & 4.88 & 104.92 & 4.58 & 111.79 & 4.57 & 112.04 & 4.61 & 111.06 & 4.71 & 108.87 \\
SP-MoE          & 4.75 & 116.73 & 4.46 & 111.34 & 4.87 & 55.78 & 4.98 & 86.93 & 4.83 & 76.16 & 4.38 & 65.24 & 4.55 & 34.62 & 4.69 & 78.11 \\
Fate            & 7.93 & 73.65 & 7.89 & 66.89 & 7.82 & 39.42 & 7.82 & 60.24 & 7.42 & 54.83 & 7.83 & 40.78 & 7.83 & 24.23 & 7.79 & 51.43 \\
HybriMoE        & 12.22 & 47.62 & 12.03 & 43.47 & 12.38 & 24.83 & 11.96 & 38.18 & 12.29 & 32.46 & 12.37 & 25.11 & 11.83 & 15.45 & 12.15 & 32.45 \\
\midrule
\rowcolor{red!15} \ProjectName{} (Ours)
 & \textbf{25.80} & \textbf{23.11} 
 & \textbf{24.10} & \textbf{22.26} 
 & \textbf{27.82} & \textbf{11.34} 
 & \textbf{28.24} & \textbf{16.93} 
 & \textbf{23.15} & \textbf{17.88} 
 & \textbf{22.78} & \textbf{14.31}
 & \textbf{23.56} & \textbf{8.36}
 & \textbf{25.06} & \textbf{16.31} \\
 \textit{Rel.Imprv.}
 & \textit{49.2\%} & \textit{29.7\%} 
 & \textit{37.1\%} & \textit{23.3\%} 
 & \textit{46.2\%} & \textit{24.8\%} 
 & \textit{50.4\%} & \textit{29.5\%} 
 & \textit{28.5\%} & \textit{16.9\%} 
 & \textit{38.4\%} & \textit{22.1\%} 
 & \textit{42.9\%} & \textit{19.8\%} 
 & \textit{41.9\%} & \textit{24.6\%} \\
\bottomrule
\end{tabular}
}
\end{table*}
Table~\ref{tab:main_result} shows that \ProjectName{} improves average TPS by \textbf{4.04$\times$} over the original baselines and by \textbf{41.9\%} over the strongest baseline, \texttt{llama.cpp-w/SD}. The latter isolates the benefit of heterogeneous expert scheduling beyond SD alone.

\subsection{Precision of Speculative Utility Estimator}

\begin{wrapfigure}{r}{0.5\textwidth} 
\centering

\includegraphics[width=0.48\textwidth]{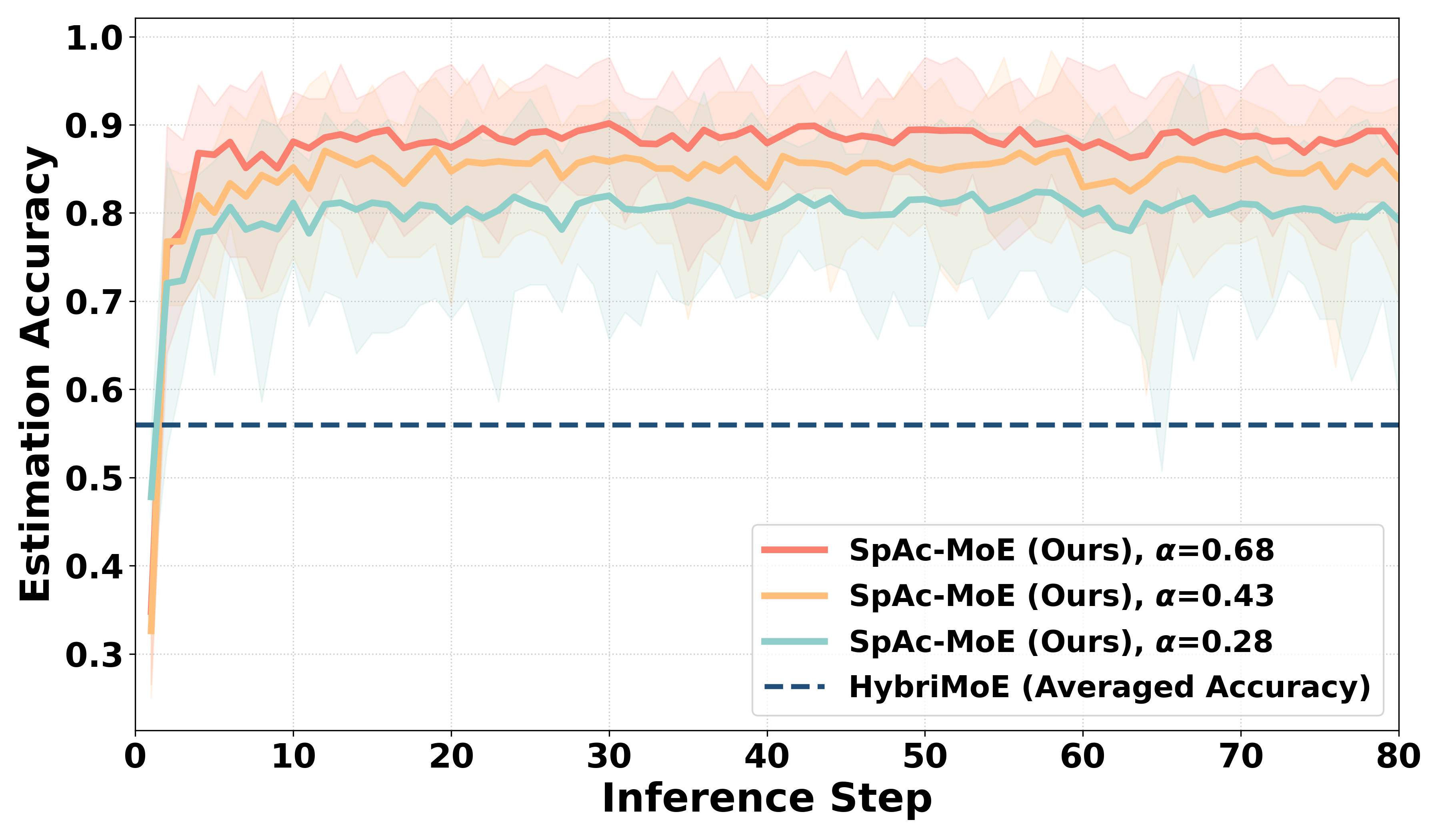}
\caption{The hot-or-cold online prediction accuracy of \ProjectName{} (SD, step level) with different acceptance rate and HybriMoE (AR, layer level).}
\label{fig:online}
\vspace{-2em}
\end{wrapfigure}

\lsh{To demonstrate the effectiveness of regarding SD as a lookahead sensor for activation utility, we report the online hot-or-cold prediction accuracy of Speculative Utility Estimation from \ProjectName{} (SD), predicted at every step under different acceptance rates, and that of HybriMoE (AR), predicted at the layer level, in Figure~\ref{fig:online}. For fairness, we report our precision averaged across all layers with the threshold $\tau$ set to 1.}

We observe that \ProjectName{} can rapidly learn activation patterns and consistently achieve a prediction accuracy of around 0.87, outperforming AR-based utility modeling approaches such as HybriMoE.

\lsh{Furthermore, we conduct experiments using less capable draft models (\texttt{Qwen3-1.7B} and \texttt{Qwen3-1.7B-int4}), which yield lower acceptance rates, and the precision correspondingly declines. This supports the following conclusion: ``The better the alignment between the draft model and the target model, the more effective SD is as a lookahead sensor.''}

\textbf{Frequency stability.} Lower sMAPE accompanies higher prediction accuracy: as sMAPE decreases from 0.889 to 0.322, accuracy rises from 0.770 to 0.897. Appendix~\ref{app:design rationale utility estimation} gives the metric definition and cross-model measurements.

\subsection{Ablation Study} \label{sec:ablation}

To assess the impact of each component in \ProjectName{}, we conduct the ablation study on MMLU-Pro, HumanEval, and CNN/DM benchmarks. 
We define the variants as follows: 
\textbf{(1)} w/o \texttt{SpecUE} removes speculative utility estimator, falling back to binary utility modeling with $K=1$;
\textbf{(2)} w/o \texttt{AdaBC} removes adaptive boundary calibration and uses fixed boundaries ($\theta^{\uparrow}_{i,t}=3$, $\theta^{\downarrow}_{i,t}=1$); 
\textbf{(3)} w/o \texttt{HetWB} removes heterogeneous workload balancer and employs a static threshold ($\tau=2$) instead of solving the online optimization; 
\textbf{(4)} w/o \texttt{SG-Prefetcher} removes utility-guided prefetcher, falling back to static expert allocation; 
\textbf{(5)} w/o \texttt{SG-Evictor} removes utility-guided evictor by retaining experts until VRAM is full; 
\textbf{(6)} w/o \texttt{SD} removes speculative decoding, reverting to standard autoregressive inference while retaining utility estimation over a sliding window of 10 tokens and heterogeneous scheduling.
The results are summarized in Table~\ref{tab:ablation}, and \lsh{the detailed analysis is attached in Appendix~\ref{app:ablation}}.

\begin{wraptable}{r}{0.6\textwidth}
\centering
\caption{Ablation study of \ProjectName{}.}
\label{tab:ablation}
\resizebox{0.6\textwidth}{!}{
\begin{tabular}{lcccccc}
\toprule
\multirow{2}{*}{\textbf{Model Variants}} & \multicolumn{2}{c}{\textbf{MMLU-Pro}} & \multicolumn{2}{c}{\textbf{HumanEval}} & \multicolumn{2}{c}{\textbf{CNN/DM}} \\
\cmidrule(lr){2-3} \cmidrule(lr){4-5} \cmidrule(lr){6-7}
 & TPS$\uparrow$ & Lat.$\downarrow$ & TPS$\uparrow$ & Lat.$\downarrow$ & TPS$\uparrow$ & Lat.$\downarrow$ \\
\midrule
\ProjectName{} (Full) & \textbf{25.80} & \textbf{23.11} & \textbf{28.24} & \textbf{16.93} & \textbf{22.78} & \textbf{14.31} \\
\quad w/o SpecUE & 19.53 & 29.48 & 20.56 & 25.27 & 18.43 & 19.53 \\
\quad w/o AdaBC & 25.33 & 23.48 & 28.03 & 19.03 & 22.31 & 15.98 \\
\quad w/o HetWB & 24.32 & 24.32 & 26.95 & 19.97 & 20.16 & 17.69 \\
\quad w/o SG-Prefetcher & 17.27 & 32.91 & 18.79 & 27.44 & 16.46 & 21.23 \\
\quad w/o SG-Evictor & 18.31 & 31.23 & 20.03 & 25.72 & 17.16 & 21.01 \\
\quad w/o SD (AR Mode) & 17.25 & 32.95 & 16.43 & 30.13 & 15.97 & 22.44 \\
\bottomrule
\end{tabular}
}
\end{wraptable}

\begin{figure*}[h] 
    \centering
    \includegraphics[width=0.98\textwidth]{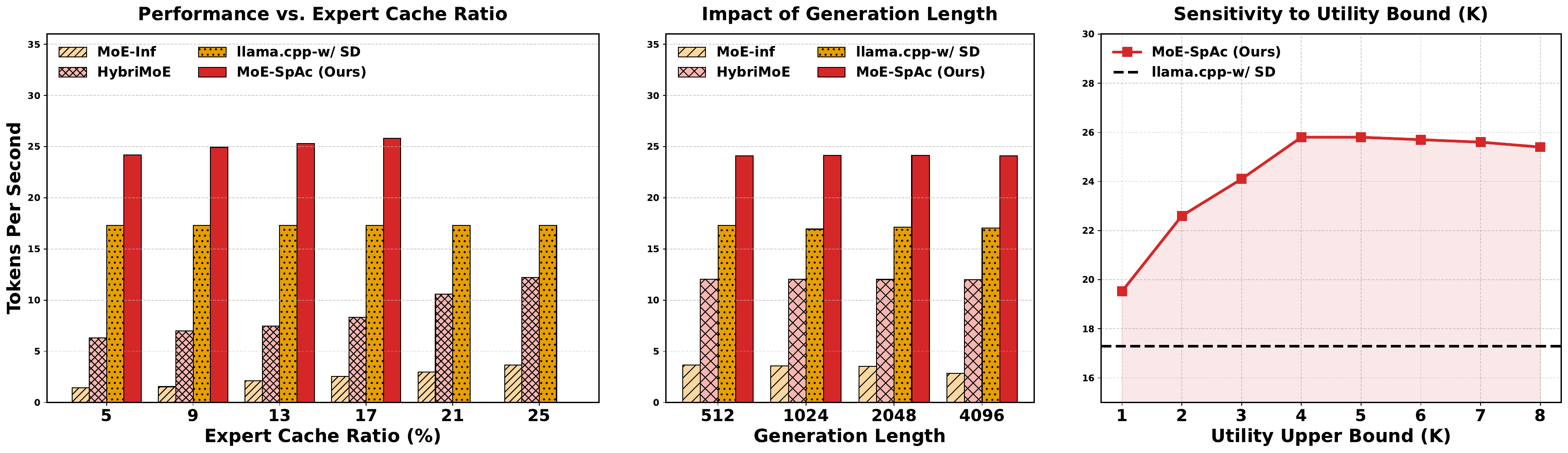}
    \caption{Stability and sensitivity analysis. Left: Impact of expert cache ratios; despite an OOM boundary at 21\% due to draft model allocation, \ProjectName{} yields superior throughput compared to existing works. Middle: Scalability across generation lengths, showing consistent gains over baselines in long-context tasks. Right: Effect of the threshold cap K, where small values reduce performance by limiting the precision of the activation utility.} \label{fig:in-depth analysis}
\end{figure*}
\subsection{In-Depth Analysis}
\label{sec:in-depth analysis exp}
In Figure~\ref{fig:in-depth analysis}, we analyze the sensitivity and stability of \ProjectName{} to variations in critical system resources and configurations, including expert cache ratio (Left), generation length (Mid), utility upper bound (Right).

    
\textbf{Expert Cache Ratio.}
As GPU cache capacity decreases, the workload balancer raises $\tau$ to satisfy Eq.~\ref{eq:vram_const}, assigning more marginal experts to CPU. Despite the draft model's approximately 8\% memory overhead, \ProjectName{} at a 17\% expert-cache ratio outperforms baselines with 25\% (Figure~\ref{fig:in-depth analysis}, Left).

\textbf{Generation Length.}
The throughput advantage persists as maximum generation length increases from 512 to 4096 (Figure~\ref{fig:in-depth analysis}, Mid), supporting stable performance over longer requests.

\textbf{Utility Upper Bound $K$.}
Figure~\ref{fig:in-depth analysis} (Right) varies $K$ from 1 to 8. Values below 4 reduce scoring precision and restrict threshold selection; larger values yield little additional benefit, supporting $K=4$.

\begin{table}[h]
\vspace{-0.5em}
\centering
\caption{TPS performance comparison across different models and datasets. We compare our \texttt{MoE-SpAc} against best baselines. \textit{Rel. Improv.} is calculated against the strongest baseline.}
\label{tab:combined_tps}
\resizebox{\textwidth}{!}{
\renewcommand\arraystretch{1.2}
\begin{tabular}{l ccc ccc ccc ccc}
\toprule
\textbf{Hardware} & \multicolumn{3}{c}{\textbf{NVIDIA RTX 4090}} & \multicolumn{9}{c}{\textbf{NVIDIA A100-80GB}} \\
\cmidrule(lr){2-4} \cmidrule(lr){5-13}
\multirow{2}{*}{\textbf{Method}} 
& \multicolumn{3}{c}{\textbf{DeepSeek-V2-Lite (16B)}} 
& \multicolumn{3}{c}{\textbf{GLM-4.5-Air (106B)}} 
& \multicolumn{3}{c}{\textbf{Qwen3.5 (122B)}} 
& \multicolumn{3}{c}{\textbf{MiniMax-M2.5 (228B)}} \\
\cmidrule(lr){2-4} \cmidrule(lr){5-7} \cmidrule(lr){8-10} \cmidrule(lr){11-13}
 & MT-bench & HumanEval & CNN/DM & MT-bench & HumanEval & CNN/DM & MT-bench & HumanEval & CNN/DM & MT-bench & HumanEval & CNN/DM \\
\midrule
HybriMoE 
 & 18.43 & 17.95 & 18.43 
 & 15.34 & 15.28 & 16.83 
 & 9.25 & 9.25 & 9.27 
 & 4.47 & 4.48 & 4.44 \\
Llama.cpp (w/ SD) 
 & \underline{30.78} & \underline{30.49} & \underline{33.23} 
 & \underline{17.11} & \underline{18.73} & \underline{17.03} 
 & \underline{12.46}  & \underline{12.78} & \underline{12.44} 
 & \underline{5.35 }& \underline{5.87 }& \underline{5.37 }\\
\rowcolor{red!10} \textbf{MoE-SpAc (Ours)} 
 & \textbf{47.10} & \textbf{47.26} & \textbf{50.17} 
 & \textbf{27.47} & \textbf{27.79} & \textbf{26.98} 
 & \textbf{20.76} & \textbf{21.42} & \textbf{20.27} 
 & \textbf{10.13} & \textbf{10.89} & \textbf{10.03} \\
\textit{Rel. Improv.} 
 & \textit{+53.0\%} & \textit{+55.0\%} & \textit{+51.0\%}
 & \textit{+60.5\%} & \textit{+48.4\%} & \textit{+58.4\%}
 & \textit{+66.6\%} & \textit{+67.6\%} & \textit{+62.9\%}
 & \textit{+89.3\%} & \textit{+85.5\%} & \textit{+86.8\%} \\
\bottomrule
\end{tabular}
}
\vspace{-1em}
\end{table}
\subsection{Generalization to Batch Size, Backbone LLM, and Hardware}\label{sec:generalization}

\begin{wrapfigure}{r}{0.45\textwidth} 
    \centering
    \includegraphics[width=0.45\textwidth]{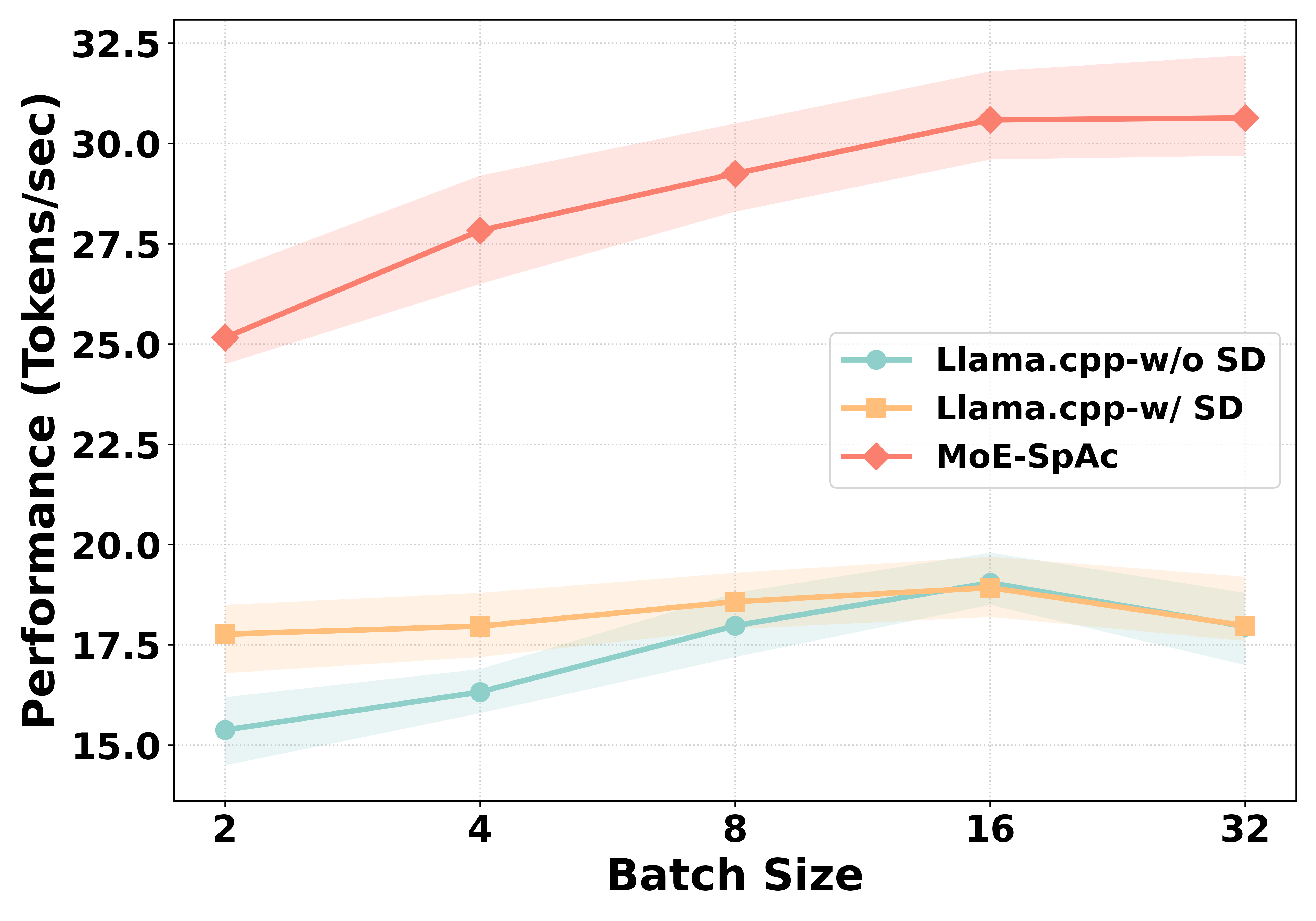}
    \caption{TPS under different batch sizes, with experiment configuration in Section~\ref{sec:setup}.}
    \label{fig:batch_size}
    \vspace{-1em}
\end{wrapfigure}

\textbf{Batch Size.} As depicted in Figure~\ref{fig:batch_size}, CPU computation, memory bandwidth, and rejected speculative work limit batch scaling; \ProjectName{} benefits from parallel CPU--GPU execution. In the additional batch-size comparison (4--32), \ProjectName{} achieves approximately twice the throughput of HybriMoE, consistent with the benefit of parallel GPU execution for hot experts. Table~\ref{tab:batch-draft} examines the joint effect of batch size and draft length.

\textbf{Compatibility to Backbone LLMs and Hardware.} We compare our method against the strongest baselines from both categories, using \texttt{DeepSeek-V2-Lite}\cite{deepseekv2}, \texttt{GLM-4.5-Air}\citep{zeng2025glm}, \texttt{Qwen3.5}\citep{qwen2026qwen35_122b_a10b}, and \texttt{MiniMax-M2.5}\citep{minimaxai2026minimaxm27}, with quantized variants serving as draft models. We further include an additional hardware setting, NVIDIA A100-80GB GPUs connected to CPU via PCIe 4.0 (32 GB/s). The results are reported in Table~\ref{tab:combined_tps}.
Across these models, average TPS improvements over the strongest baseline are \textbf{52.9\%}, \textbf{55.77\%}, \textbf{65.7\%}, and \textbf{87.2\%}, respectively.

\textbf{Smaller GPU memory.} On an RTX 4060 (8 GB), using Qwen3-1.7B to draft for Qwen3-30B-A3B, \ProjectName{} achieves 15.46 TPS versus 11.36 without SD and 11.70 with SD for llama.cpp, a 32.1\% gain over the SD baseline.

\textbf{Joint Effect of Batch Size and Draft Length.}
Table~\ref{tab:batch-draft} shows that $\gamma=8$ performs best across batch sizes 1--32. Larger batches increase memory traffic, while rejected drafts add verification work; the optimal pairing depends on acceptance, expert reuse, and memory capacity. At $\gamma=8$, increasing the batch size from 1 to 32 improves throughput by 29.3\%, indicating sublinear scaling. Extending the draft from 8 to 10 tokens yields no gain at any tested batch size; at batch size 32, the difference is only 0.4\%. This pattern favors a moderate speculative window that amortizes weight movement without excessive verification work.
\begin{table}[H]
\centering
\small
\vspace{-1em}
\caption{\ProjectName{} throughput (TPS) across batch size and draft length $\gamma$. It shows the best combination of batch sizes and draft lengths in practice.}
\label{tab:batch-draft}
\begin{tabular}{rrrrrr}
\toprule
Batch size & $\gamma=2$ & $\gamma=4$ & $\gamma=6$ & $\gamma=8$ & $\gamma=10$ \\
\midrule
1 & 20.79 & 22.23 & 23.92 & {24.10} & 23.88 \\
2 & 20.67 & 23.42 & 24.13 & {26.01} & 24.93 \\
4 & 23.83 & 23.94 & 25.78 & {27.52} & 26.63 \\
8 & 24.62 & 24.89 & 28.01 & {29.23} & 28.44 \\
16 & 25.37 & 26.41 & 29.39 & {30.48} & 29.31 \\
32 & 27.50 & 27.22 & 29.64 & {31.15} & 31.03 \\
\bottomrule
\end{tabular}
\end{table}

\section{Conclusion}
We present \textit{\ProjectName{}}, which repurposes speculative decoding as a lookahead signal for utility-guided heterogeneous expert scheduling in memory-constrained edge environments. By combining speculative utility estimation, dynamic CPU--GPU workload balancing, and unified asynchronous prefetching and eviction, \ProjectName{} reduces memory-transfer bottlenecks and improves resource utilization. Across seven benchmarks, it delivers a 42\% throughput gain over the strongest SD-based baseline and a 4.04$\times$ average speedup over standard baselines, with additional evaluations demonstrating applicability across model backbones and hardware configurations.

Our scheduling approach is complementary to advances in speculative decoding. KV-cache compression~\citep{sun2024triforce} and improved draft--target alignment~\citep{li2024eagle} offer opportunities to further increase prediction accuracy. Self-speculation~\citep{zhang2023draft} can eliminate a separate draft model, freeing memory for expert caching. Another natural integration path is provided with built-in multi-token prediction (MTP) from frontier models including GLM~\citep{glm}, DeepSeek~\citep{deepseek}, Kimi~\citep{kimi}, and Qwen~\citep{qwen}. Beyond conventional MoE models, architectures such as MoLE~\citep{jie2025mixture} and Engram~\citep{Cheng2026ConditionalMemory} introduce additional forms of sparsity and structured memory access, creating opportunities for utility estimation and proactive data placement. As sparse architectures evolve, anticipating which parameters and memory resources will be needed next offers a promising direction for efficient local inference.


\begin{ack}
Lin's research is partially supported by the National Natural Science Foundation of China (NSFC)
[Grant NSFC-624B2096, 72542012, 72595872].
Ge's research is partially supported by the National Natural Science Foundation of China (NSFC) [Grant NSFC-72225009, 72688301, 72394360, 72394365].
\end{ack}


\clearpage
\bibliography{example_paper}
\bibliographystyle{IEEEtranN}


\newpage

\appendix

\begin{tcolorbox}[
    colback=gray!10,     
    colframe=gray!50,    
    arc=4pt,             
    boxrule=0.5pt        
]
    \startcontents[appendices]
    \printcontents[appendices]{}{0}{\section*{Appendix Contents}}
\end{tcolorbox}

\section{Theoretical Analysis on Advantages of SD for MoE Inference}
\label{app:theretical analysis on SD}
We analyze the theoretical advantages of speculative decoding (SD). We demonstrate that beyond simple speedup, SD fundamentally alters the nature of the online heterogeneous expert scheduling problem by providing \textit{Expert Reuse}, \textit{Information Gain}, and \textit{Fault Tolerance}.

\subsection{Expert Reuse}\label{app:expert_resuse}

\begin{figure}[h]
\centering
\includegraphics[width=0.68\textwidth]{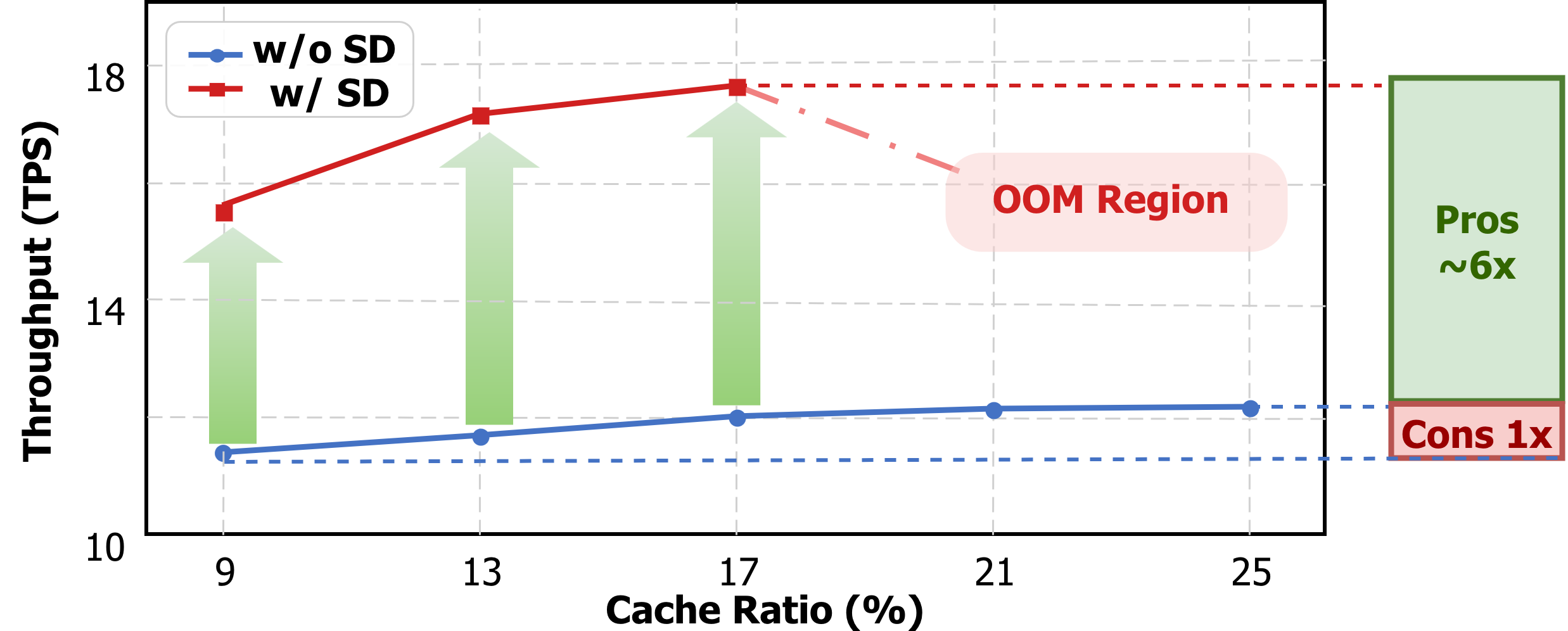}
\caption{Illustration of TPS metric to the expert cache ratio(\%) on NVIDIA GeForce RTX 4090 (24GB). Target model is \texttt{Qwen3-30B-A3B}, and draft one is \texttt{Qwen3-4B-FP8}. In edge scenario, less cache ratio precipitates only negligible decelerate, shown as red block with ``Cons'' label. Placing a small draft model can lead to huge speedup, shown as the green block with ``Pros'' label. Pros is about $6\times$ overweight Cons.
}
\label{fig:ob2-mem}
\end{figure}

We first address the rationale for allocating scarce edge memory to a draft model rather than caching more experts. 
Figure~\ref{fig:ob2-mem} illustrates the Tokens Per Second (TPS) performance of \texttt{Qwen3} models w.r.t. the percentage of total expert weights resident in GPU memory.
The result reveals that marginally increasing the resident expert set yields diminishing returns in throughput for memory-bound regimes. 
Conversely, reserving a small partition of VRAM for a draft model (8\% compared to all expert weights) enables SD, which unlocks substantial parallelism and expert reuse.

Although some works suggest SD is inefficient for MoE due to dynamic expert activation across different draft tokens~\citep{huang2025moesd, li2024eagle}, we show that when $\gamma$ (draft length) is tuned correctly, the \emph{expert reuse} across the speculation window amortizes the loading cost.
Aside from the original improvement in wall time, we now illustrate the scenario where the expert reuse in resource-constrained MoE architecture can achieve speedup.

Formally, following the notions in Section~\ref{sec:prelim_sd}, $T_V$ is dominated by FFN calculation, which is presented as 
\begin{equation}
T = (k \cdot N) \cdot (T_{LOAD} + T_{FFN}) \cdot L,
\end{equation}
where $k$ is the coefficient of the proportion of the activated experts, $N$ is the total number of experts needed for $\gamma$ tokens, $T_{LOAD}$ is the time for loading one expert, $T_{FFN}$ is the time for calculating through one expert, and $L$ is the the number of layers.

Because of the \emph{expert reuse}, $k$s for SD and AR are hard to describe with unified variable (not linear), thus denoting them as $a$ and $b$, respectively. 
Therefore, we have
\begin{equation}
\begin{aligned}
T_{SD} &= (a \cdot N) \cdot (T_{LOAD} + T_{FFN}) \cdot L + T_D \cdot \gamma \\
T_{AR} &= (b \cdot N) \cdot (T_{LOAD} + T_{FFN}) \cdot L \cdot \gamma
\end{aligned}\label{eqa:time_elapsed}
\end{equation}

\begin{figure}[h]
\centering
\begin{subfigure}{0.38\textwidth}
    \centering
    \begin{tabular}{ll|ll}
        \hline
        \textbf{$\gamma$} & \textbf{$\frac{a}{b}$} & \textbf{$\gamma$} & \textbf{$\frac{a}{b}$} \\
        \hline
        1  & 1.57 & 8  & 4.00 \\
        2  & 2.05 & 9  & 4.13 \\
        3  & 2.64 & 10 & 4.30 \\
        4  & 2.80 & 11 & 4.51 \\
        5  & 3.16 & 12 & 4.69 \\
        6  & 3.36 & 13 & 4.82 \\
        7  & 3.68 & 14 & 5.00 \\
        \hline
    \end{tabular}
    \vspace{0.5cm}
    \caption{Table on experimental data points.}
    \label{tab:math}
\end{subfigure}
\begin{subfigure}{0.48\textwidth}
    \centering
    \includegraphics[width=\linewidth]{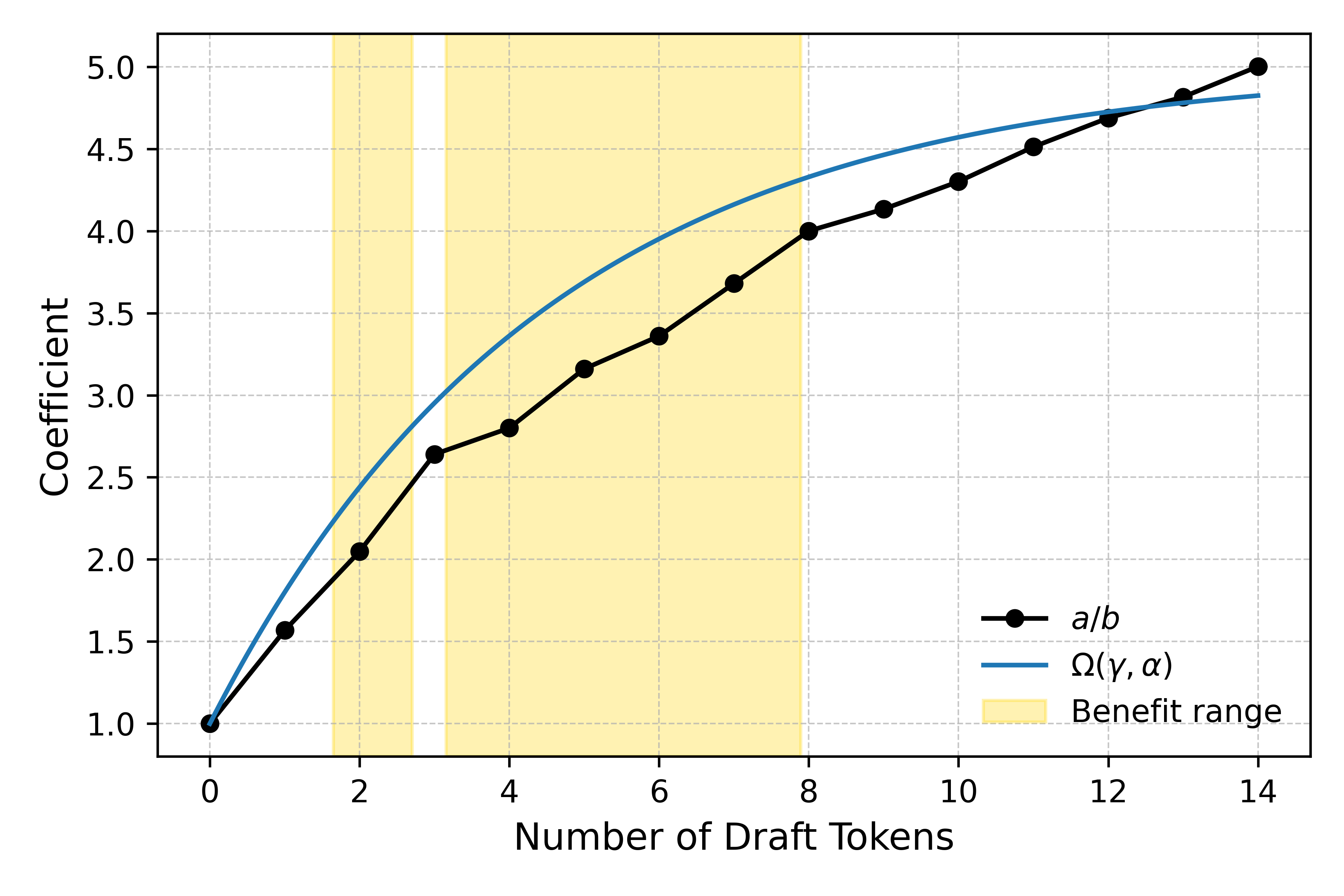}
    \caption{Experimental vs theoretical curve.}
    \label{fig:math}
\end{subfigure}
\caption{The curve comparison between $\Omega(\gamma, \alpha)$ and $\frac{a}{b}$, experimented on MT-bench with \texttt{Qwen3-235B-A22B} as verification model and \texttt{Qwen3-4B-FP8}, taking $\alpha$ as $0.8$. (a) shows the numerical result, while (b) illustrates the benefit area.}
\label{fig:math_all}
\end{figure}

For $N \cdot (T_{LOAD} + T_{FFN}) \cdot L$, which is constant, let us denote $Z$. Then the Time Per Output Token (TPOT) for both SD and AR is:
\begin{equation}
\begin{aligned}
TPOT_{SD} &= \frac{Z\cdot a + \gamma \cdot T_D}{\Omega(\gamma,\alpha)}\\
TPOT_{AR} &= \frac{Z \cdot b \cdot \gamma}{\gamma} = Z \cdot b
\end{aligned}
\end{equation}

We next describe under which scenario, $TPOT_{SD} < TPOT_{AR}$ will be established, i.e.
\begin{equation}
\begin{aligned}
    a + \gamma \cdot \frac{T_D}{Z} &< b \cdot \Omega(\gamma,\alpha)
\end{aligned}
\end{equation}

Since $\gamma \cdot \frac{T_D}{Z}$ is relatively small, it suffices to guaranteed: 
\begin{equation}
    \frac{a}{b} < \Omega(\gamma,\alpha)
\end{equation}

As Figure~\ref{fig:math_all} illustrates for \texttt{Qwen3} series, the advantage becomes more \emph{pronounced} when $\gamma \in {2} \cup [3, 8]$. Thus, expert reuse benefits inference speed in theory and provide a guideline for setting the hyper-parameters.

Notice that $\Omega(\gamma,\alpha)$ has the limit:
\begin{equation}\label{for:lim_of_omega}
    \lim_{\gamma\to+\infty}\Omega(\gamma,\alpha)=\frac{1-0}{1-\alpha}=\frac{1}{1-\alpha}
\end{equation}
which bounds the value of $\frac{a}{b}$ if an ideal speedup is desired.  

Therefore, acceptance rate $\alpha$ is closely related to the alignment between draft model and target model. A series of models which share the same pretraining data are ideal for off-shelf model speculation usage.

\begin{tcolorbox}[colframe=gray!50, colback=gray!10, coltitle=black, sharp corners=south]
\lsh{In the end-to-end experiments, }we report the key metric $\alpha$ and TPS performance across different datasets, using Equation~\ref{for:lim_of_omega} as a bridge to demonstrate their positive correlation. Details of the experimental settings are provided in Section~\ref{sec:setup}.

\vspace{1em}
\resizebox{\textwidth}{!}{ 
\renewcommand\arraystretch{1.2} 
\begin{tabular}{ccccc}
\toprule
\textbf{Metric} & \textbf{CNN/DM} & \textbf{MT-Bench} & \textbf{GSM8K} & \textbf{HumanEval} \\
\midrule
\textbf{Acceptance Rate ($\alpha$)} & 0.56 & 0.68 & 0.83 & 0.85 \\
\textbf{$\Omega$ (Token yield) - $a/b$ (Expert cost)} & 0.23 & 0.32 & 0.64 & 0.76 \\
\textbf{MoE-SpAc TPS} & \textbf{22.78} & \textbf{24.10} & \textbf{27.82} & \textbf{28.24} \\
\bottomrule
\end{tabular}
}

\end{tcolorbox}

\subsection{Information Gain}\label{sec:info_gain}

In Section~\ref{sec:task formulation}, we illustrate how SD transforms the expert activation signals from binary indicators to frequency values. 
Here, we quantitatively demonstrate the information-theoretic advantage of this transformation.

Let $E_i(\cdot)$ be an expert with activation probability $p_i$.
In standard AR inference, the expert activation for a single step is a random variable following a Bernoulli distribution: $X_i^{(1)} \sim \text{Bern}(p_i)$.
In SD with a draft length $\gamma$, the activation generalizes to the aggregated count over the verification window, $X_i^{(\gamma+1)} = \sum_{t=1}^{\gamma+1} X_{i,t}^{(1)}$. By assuming weak inter-token correlation, this follows a Binomial distribution: $X_i^{(\gamma+1)} \sim \text{Bin}(\gamma+1, p_i)$.

\textbf{Entropy Dominance.} The Shannon entropy of the aggregated signal $X_i^{(\gamma+1)}$ strictly dominates that of $X_i^{(1)}$:
\begin{equation}
    H(X_i^{(\gamma+1)}) = - \sum\nolimits_{k=0}^{\gamma+1} p_k^{(\gamma+1)} \log p_k^{(\gamma+1)} \geq H(X_i^{(1)})
\end{equation}
where $p_k^{(\gamma+1)} = \Pr(X_i^{(\gamma+1)} = k)$. This implies that $X_i^{(\gamma+1)}$ carries strictly more bits of information regarding expert demand, reducing uncertainty for utility estimation.

\textbf{Signal-to-Noise Ratio (SNR).} 
The stability of the offloading signal is critical. 
Following \citet{dietrich2007measurement}, we define SNR as the ratio of the mean demand to its standard deviation ($\mu/\sigma$).
Then, we show that the SNR of expert activation signals brought by SD is $\sqrt{\gamma+1}$ times more robust compared with that of standard AR decoding:
\begin{equation}
\begin{aligned}
    \text{SNR}_{AR} &= \frac{p_i}{\sqrt{p_i(1-p_i)}},\\
    \text{SNR}_{SD} &= \frac{(\gamma+1) p_i}{\sqrt{(\gamma+1) p_i(1-p_i)}} = \sqrt{\gamma+1} \cdot \text{SNR}_{AR}
\label{eq:snr}
\end{aligned}
\end{equation}
The relative noise in the SD signal decreases by a factor of $\sqrt{\gamma+1}$. 
This mathematical stabilization allows for reliable expert activation scoring, whereas AR is dominated by stochastic noise, especially when $p_i \ll 1$.

\subsection{Fault Tolerance}
\label{sec:fault_tolerance}

Finally, we demonstrate that the transition from AR to SD relaxes the precision requirements of utility estimation with larger safety margins, granting the system fault tolerance.

\textbf{Scheduling Fault.} 
Recall from Section~\ref{sec:task formulation} that the scheduling decision depends on whether the estimated utility $\hat{s}_{i,t+1}$ exceeds the threshold $\tau_t$. 
We define a \textit{scheduling fault} $\mathcal{E}_i$ as a discrepancy between the ground-truth optimal decision and the system's derived decision:
\begin{equation}
    \mathcal{E}_i = \mathbb{I}\left( \mathbb{I}(s_{i,t+1} \ge \tau_t) \ne \mathbb{I}(\hat{s}_{i,t+1} \ge \tau_t) \right),
\end{equation}
Specifically, a fault corresponds to either a cache miss (False Negative) or memory waste (False Positive).

\textbf{Safety Margin Analysis.}
We analyze the fault tolerance in frequency domain by defining $\tau_t^*$ as the frequency threshold, such that $G(f_{i,t+1}) \ge \tau_t \iff f_{i,t+1} \ge \tau_t^*$. 
Hence, a scheduling fault is avoided if the estimation error does not exceed the distance between the ground truth and the threshold, which is defined as the \textit{safety margin} $\Delta_{i,t}$:
\begin{equation}
     |\hat{f}_{i,t+1} - f_{i,t+1}| \leq \underbrace{|f_{i,t+1} - \tau_t^*|}_{\text{Safety Margin } \Delta_{i,t}} \Rightarrow \mathcal{E}_i=0,
\end{equation}
where $\hat{f}_{i,t+1}$ is the estimated frequency implied by the score.

\textbf{AR Rigidity.} 
In AR with $f_{i,t+1} \in \{0, 1\}$, the threshold typically lies in the interval $\tau_t^* \in (0, 1)$. 
Consequently, the safety margin is strictly bounded by $\Delta_{i,t} < 1$. 
Since frequency counts are integers, any prediction error implies $|\hat{f}_{i,t+1} - f_{i,t+1}| \ge 1 > \Delta_{i,t}$, inevitably causing a scheduling fault (e.g., predicting 0 when truth is 1).
    
\textbf{SD Tolerance.} 
In SD with $f_{i,t+1} \in [0,\gamma+1]$, the safety margin $\Delta_{i,t}$ can be larger than 1, especially for extremely hot or cold experts.
For example, with $\gamma=5$ and $\tau_t^*=2$, an expert with demand $f_{i,t+1}=5$ has a margin $\Delta_{i,t} = 3$. 
The system is tolerant to under-estimate the frequency by 1 or 2 units (noise) without crossing the boundary $\tau_t^*$, still maintaining a correct scheduling decision ($\mathcal{E}_i = 0$).

\section{Design Rationale for Speculative Utility Estimation}
\label{app:design rationale utility estimation}
In this section, we provide the analysis from the perspective of empirical observation and theoretical justification. 
\subsection{Empirical Observation}
\begin{figure}[h]
    \centering
    \includegraphics[width=0.68\textwidth]{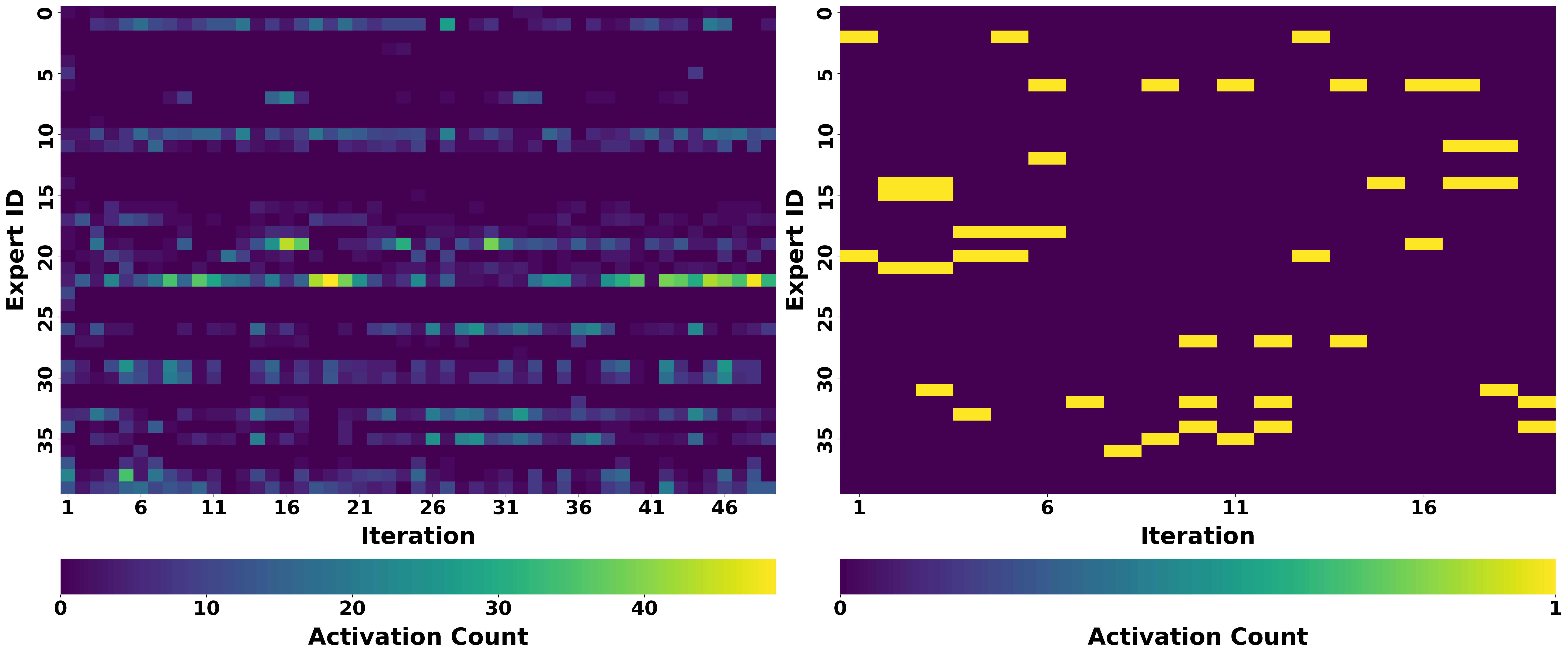}
    \caption{Activation frequency across iterations in SD is shown on the left, while the AR is shown on the right. It is much easier to estimate the light points where the experts are activated multiple times(hot experts).}
    \label{fig:heatmap}
\end{figure}
We construct a sequence of activation frequency of a same expert along the steps, e.g. $f_{1, 0}, f_{1, 2}, \cdots, f_{1, t}$ for expert No.1. This sequences of different experts in speculative inference are depicted in Figure~\ref{fig:heatmap}, with compared to those using AR. Empirically, these activation sequences in SD are more informative than AR and exhibit tighter correlation across steps. The frequency evolves gradually rather than fluctuate abruptly. 
\paragraph{Frequency stability across workloads.}
For consecutive expert frequencies, define sMAPE as $2|f_t-f_{t-1}|/(|f_t|+|f_{t-1}|)$, assigning zero when both vanish. This empirical variation measure associates lower values with more stable frequencies. The measured sMAPE/accuracy pairs are $(0.889,0.770)$, $(0.684,0.801)$, $(0.490,0.873)$, and $(0.322,0.897)$, supporting the use of temporal stability in utility estimation. Across Qwen3-30B, DeepSeek-V4-Flash-284B, and GLM-4.7-358B, sMAPE is 0.490, 0.454, and 0.164, respectively, indicating that temporal locality also occurs in larger sparse models.

\subsection{Theoretical Justification.}
The validity of the Inertial Utility Transition relies on the premise that the underlying routing signal is smooth; otherwise, an inertial tracker would suffer from excessive lag. We formalize this smoothness via the bounded drift of expert gating scores.

\begin{theorem}
(Bounded Drift of Expert Gating Scores) Let $\mathbf{h}_t^{(0)}$ and $\mathbf{h}_{t+1}^{(0)}$ be the input embeddings for two consecutive inference steps with bounded initial divergence $\delta_{in} = ||\mathbf{h}_t^{(0)} - \mathbf{h}_{t+1}^{(0)}||$. Assume the Attention and FFN modules in the routed model are $\alpha^l$-Lipschitz and $\beta^l$-Lipschitz respectively~\citep{kim2021lipschitz}. Let $W_r^l$ be the routing weight matrix at layer $l$ with spectral norm $C_R^l$.
The divergence in expert gating scores $\mathbf{s}^l$ (before Top-$k$ selection) is bounded by:
\begin{equation}
    ||\mathbf{s}_t^l - \mathbf{s}_{t+1}^l|| \leq C_R^l \cdot \delta_{in} \prod_{j=0}^{l-1} (1 + \sigma^j),
    \label{eq:gating_bound}
\end{equation}
where $\sigma^j = \beta^j(1 + \alpha^j)$ represents the layer-wise expansion factor. 
\end{theorem}

\paragraph{Preliminaries.}
Let the hidden state update rule in the $l$-th Transformer layer be:
\begin{equation}
    \mathbf{h}^{l+1} = \mathbf{h}^l + \text{FFN}^l(\mathbf{h}^l + \text{Attn}^l(\mathbf{h}^l)).
    \label{eq:transformer_update}
\end{equation}
We assume the Lipschitz conditions for the sub-modules:
\begin{align}
    ||\text{Attn}^l(\mathbf{h}) - \text{Attn}^l(\mathbf{h}')|| & \leq \alpha^l ||\mathbf{h} - \mathbf{h}'||, \\
    ||\text{FFN}^l(\mathbf{h}) - \text{FFN}^l(\mathbf{h}')|| & \leq \beta^l ||\mathbf{h} - \mathbf{h}'||.
\end{align}

\paragraph{Step 1: Layer-wise Divergence Propagation.}
Consider two input states $\mathbf{h}_t^l$ and $\mathbf{h}_{t+1}^l$ entering layer $l$. Let $\Delta^l = ||\mathbf{h}_t^l - \mathbf{h}_{t+1}^l||$.
Subtracting the update equations Eq.~\eqref{eq:transformer_update} for both states:
\begin{align}
    ||\mathbf{h}_t^{l+1} - \mathbf{h}_{t+1}^{l+1}|| &\leq ||\mathbf{h}_t^l - \mathbf{h}_{t+1}^l|| + \beta^l \left( ||\mathbf{h}_t^l - \mathbf{h}_{t+1}^l|| + ||\text{Attn}^l(\mathbf{h}_t^l) - \text{Attn}^l(\mathbf{h}_{t+1}^l)|| \right) \\
    &\leq \Delta^l + \beta^l (\Delta^l + \alpha^l \Delta^l) \\
    &= (1 + \beta^l + \alpha^l \beta^l) \Delta^l.
\end{align}
Let $\sigma^l \triangleq \beta^l (1 + \alpha^l)$. The recurrence relation becomes:
\begin{equation}
    \Delta^{l+1} \leq (1 + \sigma^l) \Delta^l.
\end{equation}

\paragraph{Step 2: Unrolling the Recursion.}
Recursively applying the inequality from layer $0$ (input embedding) to layer $l$:
\begin{equation}
    \Delta^l \leq \Delta^0 \prod_{j=0}^{l-1} (1 + \sigma^j),
\end{equation}
where $\Delta^0 = \delta_{in}$ is the initial divergence between the contexts.

\paragraph{Step 3: MoE Router Projection.}
In an MoE layer, the gating scores $\mathbf{s}^l \in \mathbb{R}^E$ (before softmax and Top-K) are computed via a linear projection $W_r^l$:
\begin{equation}
    \mathbf{s}^l = W_r^l \mathbf{h}^l.
\end{equation}
The magnitude of the difference in gating scores is bounded by the spectral norm of the weight matrix $C_R^l = ||W_r^l||_2$:
\begin{align}
    ||\mathbf{s}_t^l - \mathbf{s}_{t+1}^l|| &= ||W_r^l (\mathbf{h}_t^l - \mathbf{h}_{t+1}^l)|| \\
    &\leq ||W_r^l|| \cdot ||\mathbf{h}_t^l - \mathbf{h}_{t+1}^l|| \\
    &= C_R^l \Delta^l.
\end{align}

\paragraph{Conclusion.}
Substituting the unrolled bound for $\Delta^l$, we obtain the final inequality:
\begin{equation}\label{eq:lip_final}
    ||\mathbf{s}_t^l - \mathbf{s}_{t+1}^l|| \leq C_R^l \cdot \delta_{in} \prod_{j=0}^{l-1} (1 + \sigma^j).
\end{equation}
This confirms that the expert routing scores exist within a bounded region around the previous step's scores, scaled by the depth of the network and the Lipschitz constraints of the layers. \qed

Eq.~\ref{eq:lip_final} bounds routing-score drift, not changes in the discrete Top-$k$ set. If $\|\Delta\mathbf{s}\|_\infty\leq\epsilon$, Top-$k$ membership is preserved only when the gap between the $k$-th and $(k+1)$-th scores exceeds $2\epsilon$. Near ties, arbitrarily small perturbations can change membership. Multi-token aggregation can reduce sensitivity to individual flips empirically, but does not guarantee stable frequencies or error-free prediction. The inertial estimator is therefore motivated by the observed temporal correlation; CPU fallback handles prediction errors.

\section{Optimization Formulation \& Solution for Heterogeneous Workload Balancer}\label{pip}
\subsection{Formulation Explanation}

\begin{figure}[t]
  \centering
  \begin{subfigure}[t]{0.48\textwidth}
    \centering
    \includegraphics[width=\textwidth]{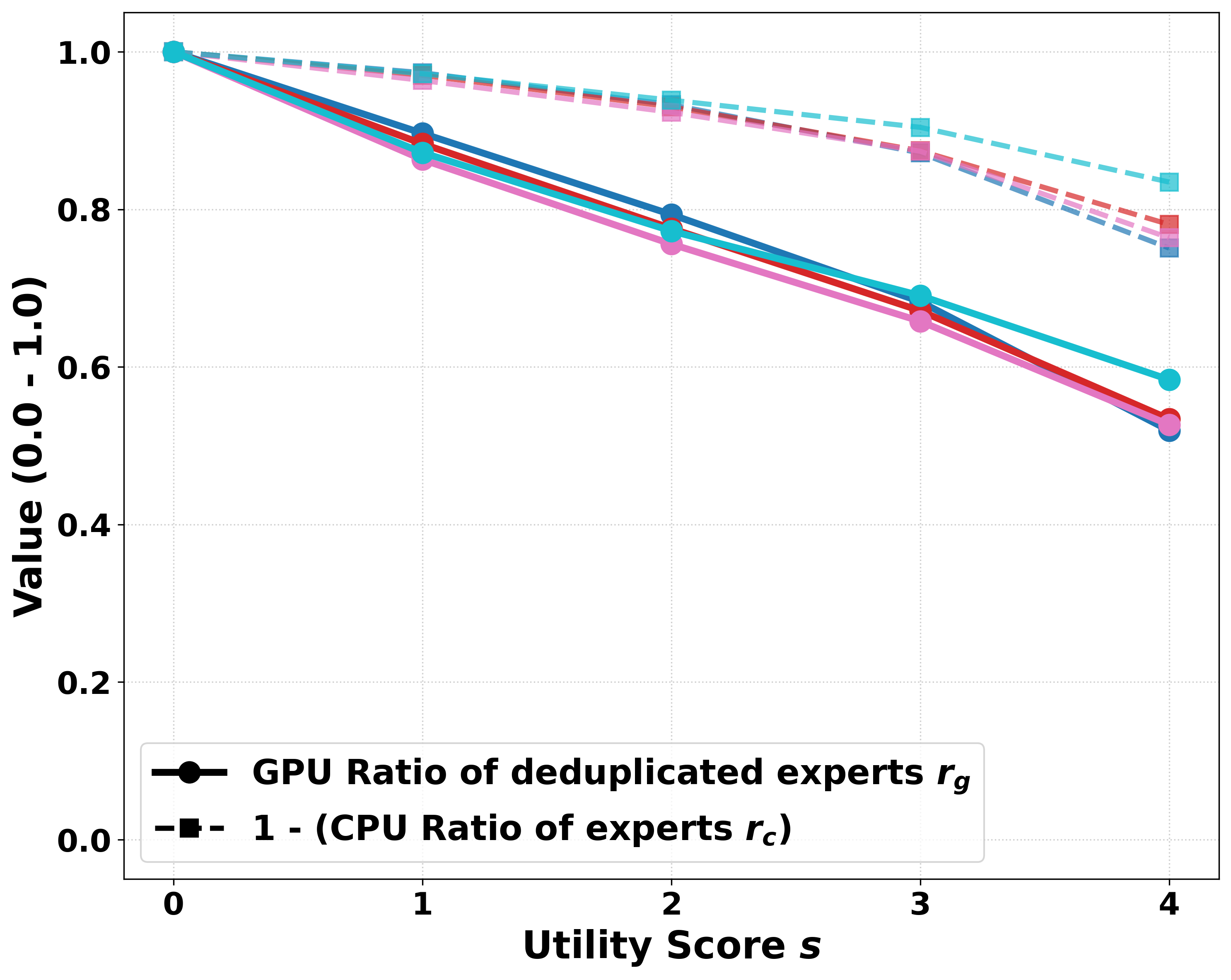}
    \caption{Utility score to the ratio of $\gamma k$ activated experts that can be processed on GPU, $1- r_c$ (Dotted line) and the ratio of b de-duplicated activated experts to be processed on GPU in parallel. $r_g$ (Solid line) for a single layer.}
    \label{fig:thre2hr-ur}
  \end{subfigure}
  \hfill
  \begin{subfigure}[t]{0.48\textwidth}
    \centering
    \includegraphics[width=\textwidth]{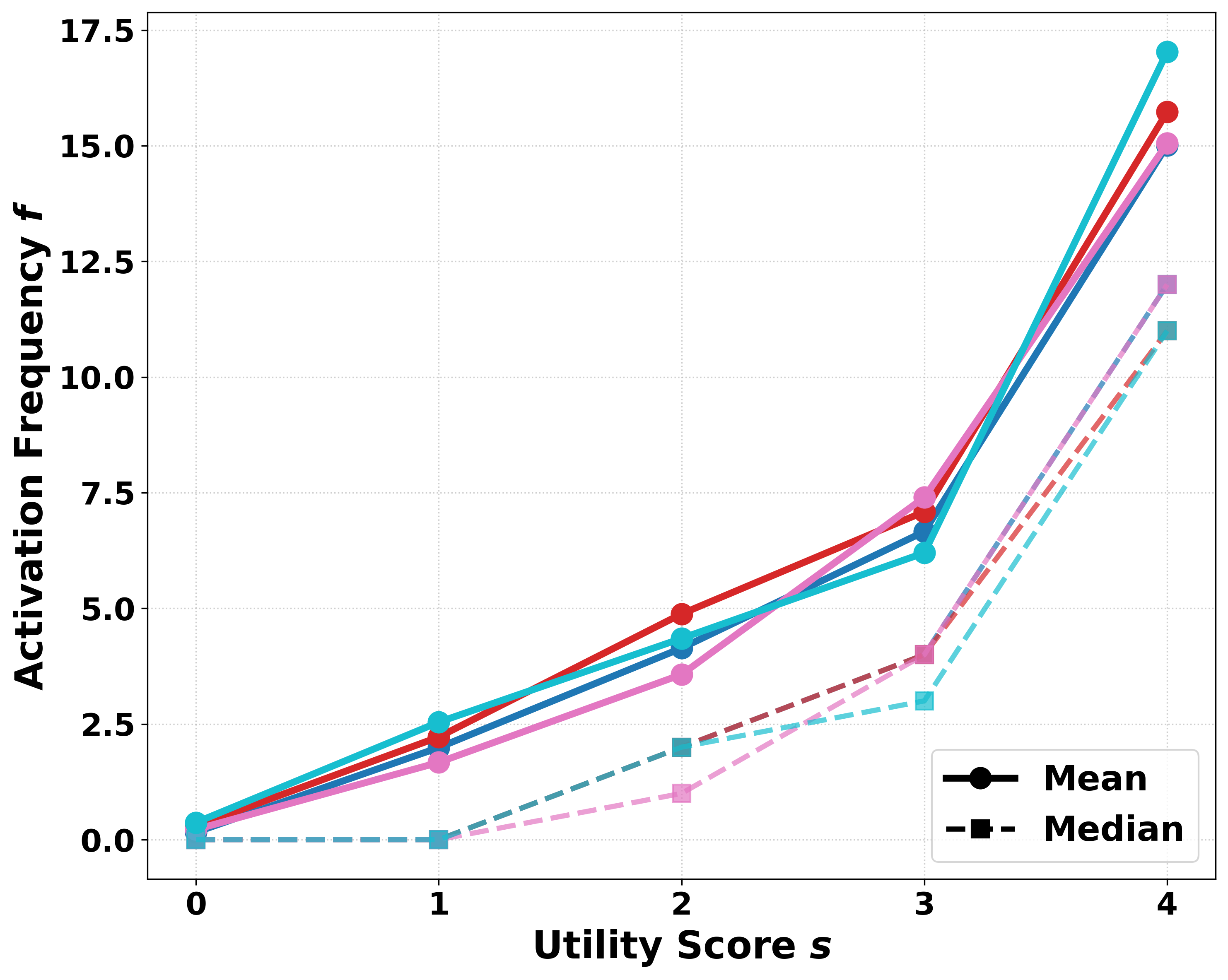}
    \caption{Utility Score to the average value (Solid line) and median (Dotted line) of activation frequency. Experiment conducted on MT-Bench with $K = 4$ on \texttt{Qwen} models.}
    \label{fig:state2fc}
  \end{subfigure}
  \caption{Relationship between utility and activated frequency, ratio of activated experts, etc..}
  \label{fig:combined}
\end{figure}

To illustrate Equation~\ref{eq:cpu_time} and Equation~\ref{eq:gpu_time}, we first look at the two functions related to $\tau$. As depicted in Figure.~\ref{fig:thre2hr-ur}, the ratio of $\gamma k$ activated experts processed on CPU $r_c(\tau)$ and the ratio of de-duplicated activated expert processed on GPU $r_g(\tau)$ remain stable in value for a given threshold $k$ within a single layer. Therefore, this value can be learned from a warm up test. 

To illustrate their relations with $T_{CPU}$ and $T_{GPU}$, say an inference step that activates two experts per token with two draft tokens ($\gamma=2$). If a layer forward pass prefetches expert No. 3 based on threshold $\tau$, while the draft tokens activate experts \{No. 0, No. 3\} and \{No. 2, No. 3\}. Then  $r_c(k) \approx 1-\frac{2}4$ and $r_g(\tau) \approx 1/3$. The host calculation time would be the sum of computations for missed experts 0 and 2, taking time $r_c(\tau) \times  4 \times T^u_{cpu}$. The device processes the stacked hidden states of the identified experts concurrently, which only takes $r_g(\tau) \times 3 \times T^u_{gpu}$.

Besides, Eq.~\ref{eq:io_const} imposes a strict constraint to prevent I/O bottlenecks. Specifically, the prefetching budget for each layer is bounded by the aggregate duration of computation and the corresponding drafting phase. By prioritizing the placement of hot experts on the limited device memory, the optimization forces the thresholding policy to align with the inter-device transmission bandwidth.
Equation~\ref{eq:integer_const} enforces a strictly positive threshold ($\tau > 0$) to differentiate our predictive approach from non-selective prefetching strategies.
Regarding Eq.~\ref{eq:vram_const}, minimizing the overhead is critical; querying the remaining memory $VRAM_{left}$ via \texttt{cudaMemGetInfo} entails non-negligible latency in CPU. Consequently, we can formulate an online linear programming problem to prioritize matching other constraints first, incorporating the memory constraint with latency-aware adjustments~\citep{agrawal2014dynamic, li2020simple}.

Furthermore, Figure~\ref{fig:state2fc} empirically demonstrates that a higher score in utility estimator correlates with a higher frequency score in computation, thereby better utilizing the high-dimensional computational capacity of the GPU~\cite{markidis2018nvidia}.

\subsection{Solution Derivation}
Since the domain of $\tau$ is restricted to integers $\tau \in [1, K]$, we determine the optimal $\tau^*$ by exploiting the structural properties of the objective function.
First, we define the feasible set $\mathcal{S}$ constrained by physical resources. Let the available computation window be $C_{window} = T_{total} + \gamma \cdot T^{u}_{draft} / L_{target}$. Based on the I/O (Eq.~\ref{eq:io_const}) and Memory (Eq.~\ref{eq:vram_const}) constraints, the feasible set is determined by the range of prefetching volume $n(\tau)$:
\begin{equation}
    \mathcal{S} = \{ \tau \in \mathbb{Z} \mid \tau_{min} \le \tau \le \tau_{max} \} \subseteq [1, K]
\end{equation}
where the bounds $\tau_{min}$ and $\tau_{max}$ are derived from the limits on $n(\tau)$ imposed by $C_{window}$ and $\text{VRAM}_{left}$.
The objective function is defined as the absolute difference between the device execution times:
\begin{equation}
    f(\tau) = \Big| \underbrace{r_c(\tau) \cdot \gamma \cdot k \cdot T^{u}_{cpu}}_{T_{cpu}(\tau)}  -  \underbrace{r_g(\tau) \cdot b \cdot T^{u}_{gpu}}_{T_{gpu}(\tau)} \Big|
\end{equation}
Since a higher threshold $\tau$ reduces the number of hot experts, the GPU allocation ratio $r_g(\tau)$ is monotonically decreasing, causing $T_{gpu}(\tau)$ to decrease. Conversely, the CPU allocation ratio $r_c(\tau)$ is monotonically increasing, causing $T_{cpu}(\tau)$ to increase. Consequently, the difference $T_{cpu}(\tau)-T_{gpu}(\tau)$ is nondecreasing, so its absolute value is unimodal, but need not be convex. A constant-time analytical crossing requires an affine time model, as in the solver microbenchmark; it is not implied by monotonicity alone. With such a model and a contiguous feasible interval, only a constant number of candidates need evaluation. The optimal solution $\tau^*$ corresponds to either the integer closest to the theoretical intersection point $\tau_{cross}$ where $T_{cpu}(\tau_{cross}) = T_{gpu}(\tau_{cross})$, or one of the boundary points of $\mathcal{S}$ if the intersection lies outside the feasible range:
\begin{equation}
    \tau^* = \operatorname*{arg\,min}_{\tau \in \{\tau_{min}, \tau_{max}, \lfloor \tau_{cross} \rfloor, \lceil \tau_{cross} \rceil\} \cap \mathcal{S}} f(\tau)
\end{equation}
The candidate rule is exact under the stated model and feasibility assumptions. For arbitrary measured ratios, exhaustive evaluation of the $K$ thresholds provides an $O(K)$ reference; a monotone crossing can be located in $O(\log K)$ when feasibility is an interval. If the feasible set is empty, prefetching must be disabled and missing experts executed on CPU; choosing $\tau=K$ alone does not ensure feasibility. These complexity bounds concern threshold selection, excluding utility updates and coefficient estimation.

\section{Implementation Details of Unified Asynchronous Execution Engine}\label{apix:impl}
\textbf{Utility-Guided Prefetching.}
To manage the prefetching requests, we implement a \textit{multi-level priority queue}, where each level contains pending requests for experts with the same utility $u_{i,t}$. Within each level, requests follow a First-In-First-Out (FIFO) order to respect the temporal sequence of layer execution for target model.
The prefetcher continuously scans the queues at different levels in descending order of utility (from $K$ down to $\tau$). 
The execution logic is filtered by the dynamic threshold $\tau_t$: the prefetcher only processes queues with levels $u_{i,t} \ge \tau_t$, transferring weights in descending order of priority (from level $K$ down to $\tau$). 

\textbf{Utility-Guided Eviction.}
Complementarily, we manage the resident GPU cache using a \textit{state-ordered pool}, implemented via a Red-Black tree. This structure indexes all GPU-resident experts by their current utility score $u_{i,t}$. 
When the load balancer updates the threshold to $\tau_t$, the evictor can efficiently identify and remove resident experts with decreased scores $u_{i,t} < \tau_t$ in $O(\log N)$ time.

Crucially, unlike traditional policies such as Adaptive Replacement Cache~\cite{megiddo2004outperforming} or Least Recently Used (LRU)~\cite{chrobak1999lru} which rely on historical access timestamps, our approach employs a \textbf{unified metric} (expert utility) for both prefetching and eviction. This unification ensures scheduling consistency, effectively preventing the cache thrashing of marginally hot experts.

\begin{figure}[h] 
\centering
\includegraphics[width=0.5\linewidth]{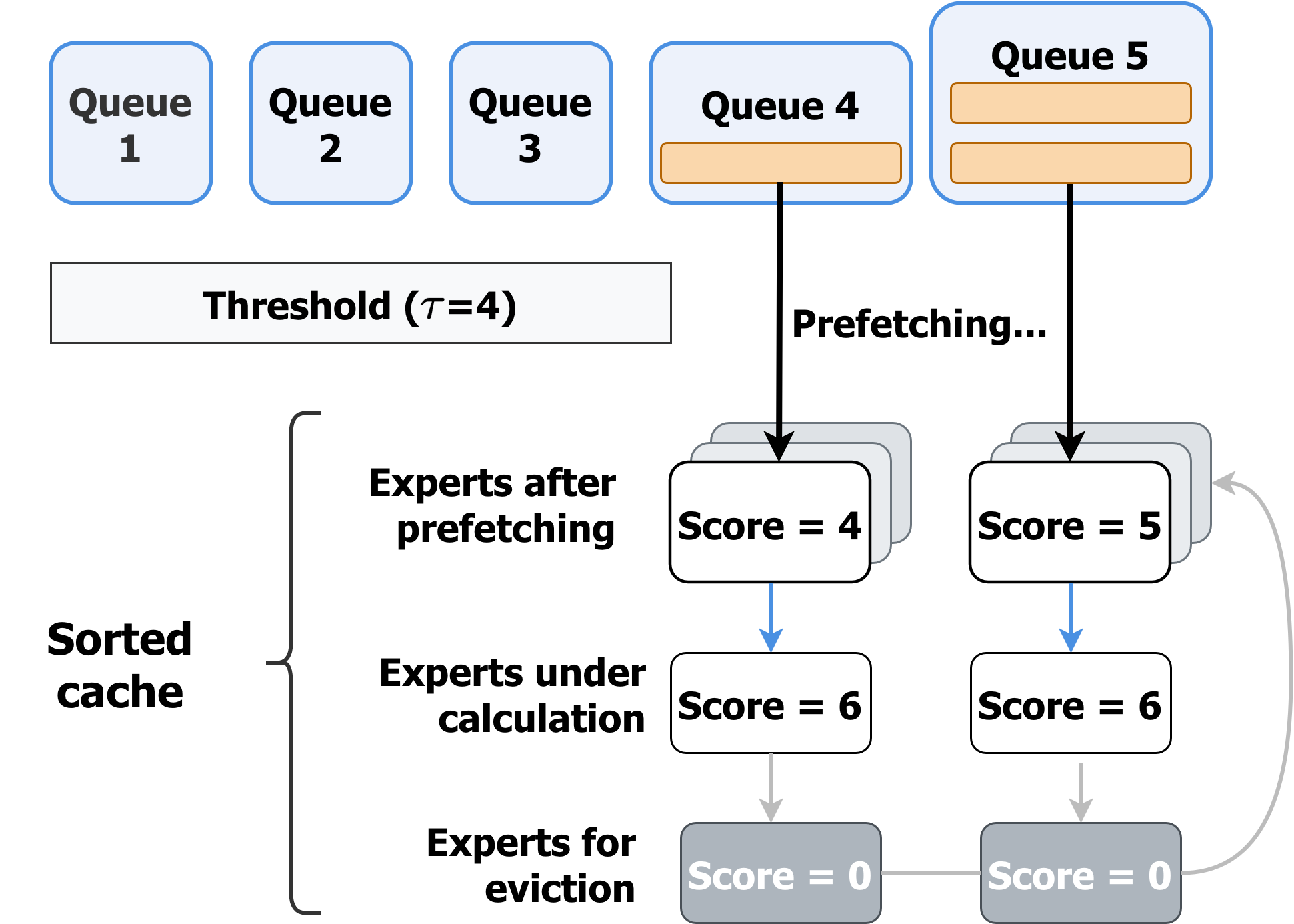}
\caption{Scheduling and cache management of the \ProjectName{} in implementation when $K=5$.} \label{fig:impl}
\end{figure}

When it comes to the management of VRAM cache and score-based prefetching method, Figure~\ref{fig:impl} shows the details. Each expert ID is pushed into the queue labeled with corresponding score. With the online decision of threshold, the queue with lower lable is filtered out. Then the experts are prefetched according to the ID reported in the queue from high to low. When prefetching is finished, each expert weight is tagged with a score and recorded in a \texttt{set}. When the expert is in used, the score is modified to $K+1$, which we call a frozen expert, because under no circumstance should it be evicted. Due to the limited resources, most experts should be evicted right after calculation, and leave the space for layers to come. Therefore, we change these scores to $0$, thus quickly recycled and refilled by other weights. All the experts weight is cache is recorded dynamically with a \texttt{set}, reaching an ideal efficiency with fine-grained lock usage.

\begin{figure}[h] 
\centering
\includegraphics[width=0.8\linewidth]{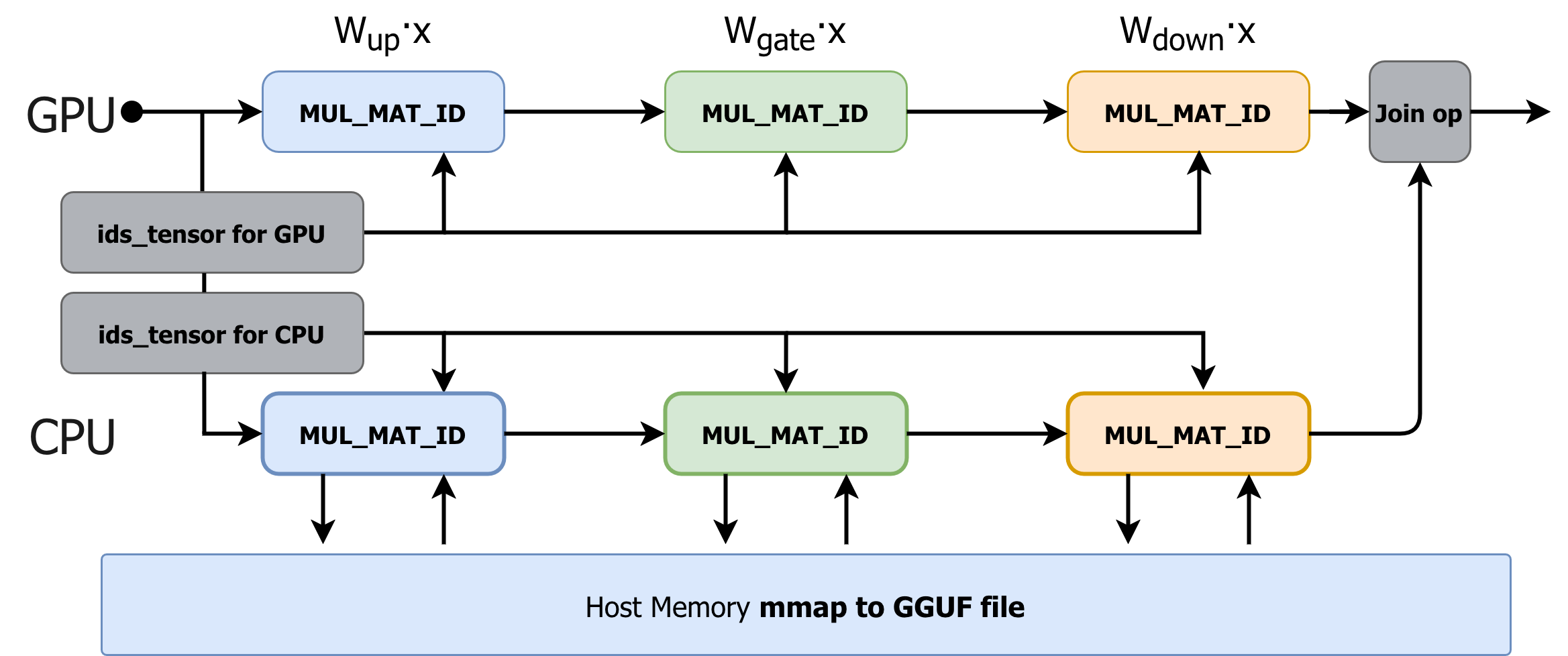}
\caption{\ProjectName{} implementation atop \texttt{llama.cpp}, unchanged operators is omitted.} \label{fig:llama-hack}
\end{figure}

We implement the management and scheduling system asynchronously atop \texttt{llama.cpp}, which native support CPU specific or GPU specific expert calculation. As shown in Figure~\ref{fig:llama-hack}, we hack a \texttt{ids\_tensor} modification in backend, rewrite the expert calculation kernel: skip the calculation of expert with special label in \texttt{ids\_tensor}. This modification is applied to each process in expert calculation in one layer forward, including $W_{up}\cdot x$, $W_{gate}\cdot x$, and $W_{down}\cdot x$. With a CUDA kernel joining up the result from the heterogeneous devices, we implement this process neatly without breaking the computation graph.

\section{Baseline Information}
\label{app:baselines}

We compare \ProjectName{} against representative inference systems covering dense and MoE models, as well as different offloading and prediction strategies as follows:
\begin{itemize}
    \item \texttt{Transformers Accelerate} \citep{wolf2019huggingface} supports layer-wise parameter offloading via device maps. 
    \item \texttt{vLLM} \citep{kwon2023efficient} is a high-throughput inference engine; our comparison enables CPU weight offloading with \texttt{--cpu-offload-gb 60}. 
    \item \texttt{llama.cpp} \citep{ollama2024} is a lightweight inference engine optimized for constrained GPU environments. 
    \item \texttt{llama.cpp-w/ SD} extends llama.cpp with speculative decoding. 
    \item \texttt{Mixtral Offloading} \citep{eliseev2023fast} offloads expert weights between CPU and GPU while maintaining an LRU cache.
    \item \texttt{MoE-Infinity} \citep{xue2024moe} is a cost-effective MoE inference library that predicts expert activation paths and evicts experts based on least visit count.  
    \item \texttt{MoE-SpeQ}~\citep{wang2025moe} uses an on-device draft model to predict future expert demand and proactively prefetch weights, with an adaptive governor tuning speculation to the hardware.
    \item \texttt{SpecExp}~\citep{madan2026speculating}, short for Speculating Experts, predicts future experts from internal model representations to overlap CPU--GPU weight transfers with computation; it also explores executing predicted experts directly.
    \item \texttt{SP-MoE} \citep{chen2025sp} predicts expert activation by directly using a draft model. 
    \item \texttt{Fate} \citep{fang2025fate} predicts activated experts using the gating output of the subsequent layer and proactively offloads expert weights to target devices.
    \item \texttt{HybriMoE}~\citep{zhong2025hybrimoe} is a hybrid CPU-GPU inference framework that improves resource utilization through a heterogeneous scheduling and cache management system, implemented atop \texttt{kTransformers}.
\end{itemize}

\section{Illustration of Inference Pipelines}

\begin{figure}[t] 
    \centering
    \includegraphics[width=0.48\textwidth]{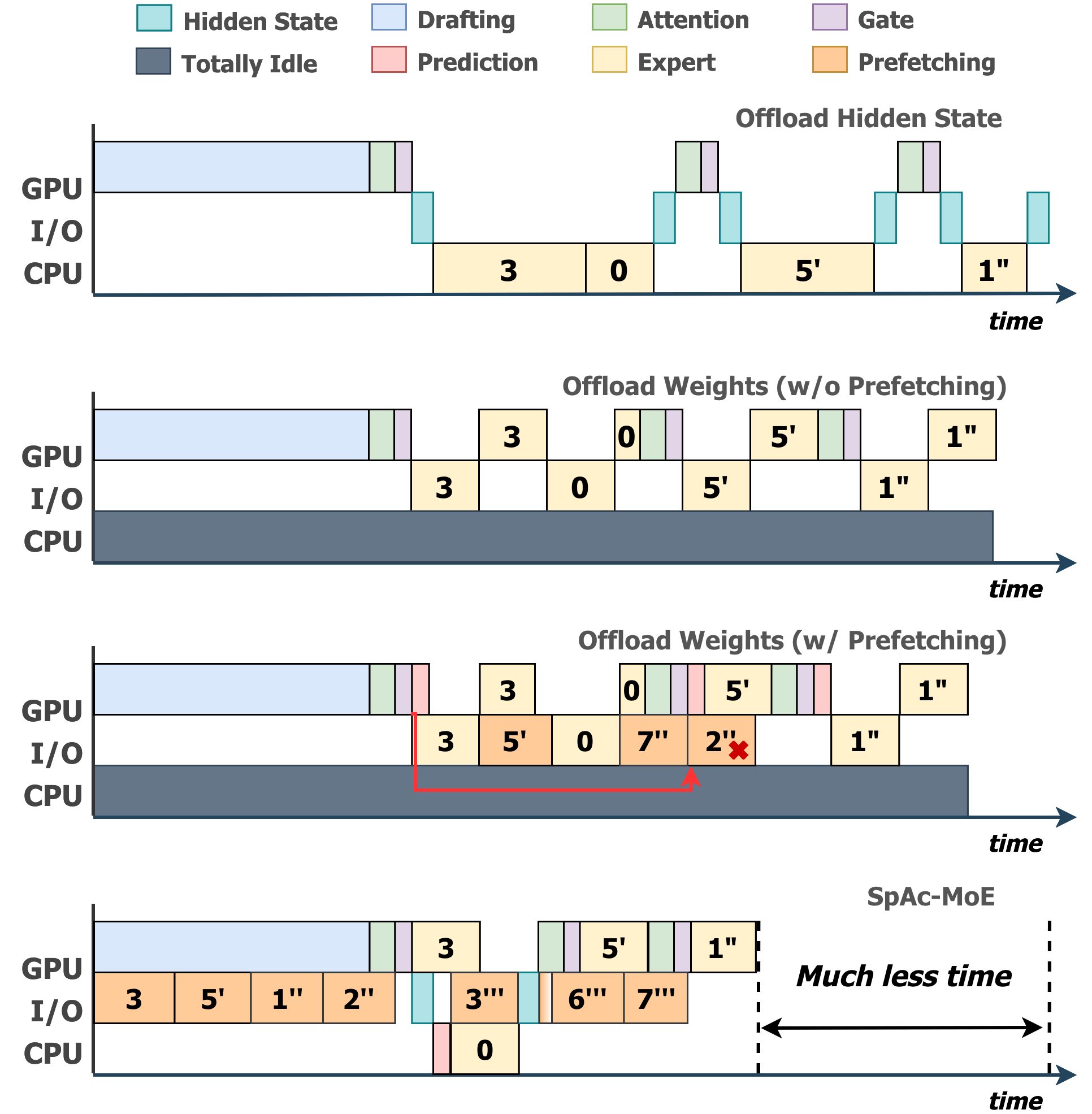}
    \caption{Comparison of MoE inference pipelines under Speculative Decoding (SD). `GPU' and `CPU' denote the compute device for expert layers, while prime notation (e.g., $1', 1''$) indicates the layer depth. \ProjectName{} (bottom) minimizes pipeline bubbles via optimized heterogeneous scheduling.} \label{fig:pip}
\end{figure}
In Figure~\ref{fig:pip}, we compare the pipeline of our proposed \ProjectName{} with other scheduling baselines. We can observe that speculative decoding with heterogeneous computation allows parallel scheduling. With load balancing from Section~\ref{sec:balancer}, we can achieve much less time elapsed compared with other expert-scheduling baselines.

\paragraph{Layer latency.}
The layer profile takes 0.73--0.78 ms for most layers with \ProjectName{}, compared with 1.26--1.35 ms for llama.cpp, where CPU computation dominates the tail. Layer 47, including the fused output head, takes 1.51 ms with a CPU computation tail of 43\%. These measurements explain the reduction in CPU bottlenecks within the verification pipeline.

\paragraph{Misprediction cost.}
A false-positive prediction consumes one expert's cache space and may incur an unnecessary weight transfer; its byte cost depends on expert dimensions and precision. A false negative executes the missing expert on CPU and transfers hidden states instead of the expert weights. CPU--GPU overlap absorbs part of this cost, while the remaining CPU work can extend the synchronization tail.

\section{Ablation Explanation}\label{app:ablation}
\lsh{In this section, we provide a detailed analysis of the ablation results in Section~\ref{sec:ablation}.}
\begin{itemize}[leftmargin=10pt]
    \item The speculative utility estimator is the cornerstone of our framework. 
    Removing it (i.e., w/o \texttt{SpecUE}) results in a significant drop in both throughput and latency, confirming that the informative frequency-valued signals brought by SD are crucial for expert demand estimation and memory management. 
    
    \item Within the estimator, the adaptive boundary calibration contributes to robustness. 
    However, its removal (w/o \texttt{AdaBC}) yields only a marginal degradation. 
    This suggests that our default parameters ($\theta^{\uparrow}=3, \theta^{\downarrow}=1$) are sufficiently general for most expert activation patterns, though adaptation refines performance for corner cases.
    
    \item The heterogeneous workload balancer proves critical for I/O management. By fixing the threshold $\tau$, the system fails to adapt to real-time I/O bandwidth constraints, leading to either I/O congestion or under-utilization of the GPU, causing a performance drop. There, we conduct more experiment with fixed $\tau$ to demonstrate the advantage of \texttt{HetWB}:

    \begin{table}[ht]
    \centering
    \caption{Performance with adaptive and fixed thresholds.}
    \resizebox{0.6\textwidth}{!}{ 
    \renewcommand\arraystretch{1.2} 
    \begin{tabular}{lcccc}
    \toprule
     & \textbf{Adaptive (Ours)} & \textbf{Fix $\tau=1$} & \textbf{Fix $\tau=2$} & \textbf{Fix $\tau=4$} \\
    \midrule
    \textbf{TPS} & \textbf{24.10} & 21.73 & 23.48 & 10.16 \\
    \bottomrule
    \end{tabular}
    }
    \end{table}

    \item The utility-guided prefetcher and evictor are tightly coupled. 
    Removing one of them (\texttt{SG-Evictor} or \texttt{SG-Prefetcher}) degrades the performance to the level of the baseline, since I/O is no longer masked. 
    This confirms that the synergy between predictive scoring, dynamic balancing, and asynchronous execution is essential for realizing the full potential of \ProjectName{}.
\end{itemize}

\paragraph{Cache hits and baseline interpretation.}
Our scheduler combines proactive GPU placement with CPU execution for cold or missed experts. GPU expert-cache hit rates are 0.71, 0.73, 0.75, and 0.70 on MT-Bench, GSM8K, HumanEval, and CNN/DM, respectively. Cache hit rate differs from hot/cold prediction accuracy: assigning activated experts to CPU can reduce the synchronization tail even when it lowers cache hits. Table~\ref{tab:main_result} includes the additional baseline comparisons; Accelerate covers on-demand weight loading with a simple cache.

\section{Implementation Validation}\label{app:revision-evidence}
\subsection{Numerical Compatibility}
Exact speculative sampling preserves the target distribution under the original acceptance/rejection rule and a numerically consistent target evaluator. Quantizing the draft changes proposal quality and acceptance. We validate the heterogeneous kernels separately to check numerical consistency between CPU and GPU computation.
\begin{table}[ht]
\centering
\caption{Numerical validation. $E$, $U$, and $T$ denote expert count, experts per token, and token count. Split-pipeline tests cover cache ratios of 0/50/100\%. NMSE is normalized mean squared error.}
\label{tab:numerical-validation}
\small
\begin{tabular}{p{0.24\linewidth}p{0.49\linewidth}r}
\toprule
Test & Configuration & Worst NMSE \\
\midrule
AGGRE (CUDA vs. CPU) & $E=256$, $U=8$, $T\in\{1,32\}$ & 0 (bit-exact) \\
Small split MoE & $E=8$, $U\in\{2,8\}$, $T\in\{1,7,32\}$ & $4.1\times10^{-6}$ \\
Qwen3-like split MoE & $E=128$, $U\in\{2,8\}$, $T\in\{1,7,32\}$ & $1.2\times10^{-6}$ \\
Realistic layer & embedding 2048, FFN 768, $E=128$, $U=8$, $T=32$ & $8.3\times10^{-7}$ \\
\bottomrule
\end{tabular}
\end{table}
AGGRE is bit-exact in the tested configurations, and all split-pipeline groups pass with worst NMSE at most $4.1\times10^{-6}$. These tests establish numerical agreement at the kernel and layer levels; exact distributional preservation remains conditional on the sampling rule and target evaluation.

\subsection{Additional Runtime Observations}
The coefficient-update microbenchmark reports 4.31 ns per layer for the warmup arithmetic; this excludes collecting the profiling measurements. The author-response comparison reports that a single-activation CPU fallback using Intel AMX is within a 50\% latency gap of GPU execution, while the AVX2 path is approximately $3\times$ slower. These relative costs depend on the processor and expert configuration. The response estimates approximately 600 MB for a false-positive expert placement in its cost example; the general cache cost is the actual expert weight size $W_e^u$, which varies across models and precisions.

\section{Related Works} \label{apix:related_work}

\subsection{Efficient MoE on Edge}

MoE has emerged as a crucial architecture for scaling LLMs, prompting extensive research into optimizing its inference efficiency. \citet{zhu2026probecobalancingcomputationcommunication} co-balances computation and communication in real time in distributed system, which overlooks the stress from edge scenarios.
Model compression techniques---including pruning~\citep{xie2024moe, lee2024stun}, quantization~\citep{frantar2023qmoe, imani2024mixture}, distillation~\citep{salinas2022knowledge, shu2024llava}, and decomposition~\citep{yang2024moe, li2023merge}---have been successfully applied to MoEs. However, similar to their application in dense models, these methods inevitably trade generation quality for acceleration.

In scenarios where precision loss is not allowed and MoEs exceed GPU memory capacity, system-level offloading becomes essential. While latency in large-batch regimes can be amortized via micro-batch pipelining~\citep{cao2025moe, zhuge2025specoffload, fang2025klotski, xu2025moe}, low-latency inference for single-batch requests remains challenging.
To address this, \emph{on-demand loading} methods, such as Lina~\citep{li2023accelerating}, ExpertFlow~\citep{he2024expertflow}, and OD-MoE~\citep{wang2025odmoeondemandexpertloading}, exclusively utilize GPUs for expert computation. If an expert is not prefetched to the GPU, it must be loaded on demand. The downside of these approaches is the high I/O overhead, which causes prolonged GPU idling. \textit{Expert Prefetching}~\citep{xue2024moe, zhong2024adapmoe, fang2025fate, gavhane2025moebeyondlearningbasedexpertactivation} leverages historical activation patterns to preload experts, overlapping I/O with computation. Complementarily, \textit{Expert Caching}~\citep{he2024expertflow, tang2024hobbit, zhong2024adapmoe} exploits the temporal locality of expert activation to retain experts in high-bandwidth memory, mitigating offloading overheads. These research works are largely GPU-intensive and therefore do not fully utilize the computational resources of heterogeneous devices.

\subsection{Speculative Decoding (SD)}

SD allows for lossless latency reduction by verifying multiple drafted tokens in parallel~\citep{leviathan2023fast, xia2022speculative,xi2025efficiency, qin2026confu}. Recently, \citet{zhang2025markov} provided a precise information-theoretic rationale for SD by quantifying the information surplus in hidden states.
However, the efficiency of SD is highly sensitive to the operational intensity. As batch size increases, the system transitions from memory-bound to compute-bound, making the verification of speculated tokens prohibitively expensive~\citep{liu2024optimizing, li2024snapkv, miao2023specinfer, sun2024triforce}.
Conversely, for personal or edge usage where batch sizes are small, inference is strictly memory-bound. In this regime, the additional arithmetic operations required for SD verification are strictly encompassed by the memory retrieval time, making SD an ideal candidate for acceleration~\citep{cai2024medusa}.

\subsection{Heterogeneous Computing}
Previous offloading techniques have primarily focused on reducing memory transfer overhead by offloading certain computations to the CPU~\citep{park2024improving}. For instance, PowerInfer~\citep{song2024powerinfer} reduces GPU memory demand by executing less frequently activated neurons on the CPU, taking advantage of skewed activation patterns. Caraserve~\citep{li2024caraserve} addresses cold-start delays in LoRA serving by utilizing CPU assistance and employing rank-aware scheduling to reduce latency. However, they do not ensure skewed activations of neurons or parameter reuse.
In the context of MoE models, techniques like Fiddler \citep{kamahori2024fiddler} and kTransformers~\citep{10.1145/3731569.3764843} extend this concept by offloading expert layer computation to the CPU during cache misses. Specifically, when an expert is not in the GPU cache, the CPU executes the corresponding expert layer instead of loading it from memory. These approaches aim to optimize memory usage by exploiting CPU-GPU parallelism and mitigating the overhead of loading large models onto the GPU, but still remains a huge waste.
HybriMoE~\citep{zhong2025hybrimoe} introduces a queue that re-orders experts by token count, dispatching hot experts to the GPU and cold ones to the CPU. It uses a greedy algorithm to minimize per-layer computation cost and prefetches experts for the next layer during non-expert computations to further reduce on-demand loading pressure. But its greedy algorithm and scoring-based cache strategy lacks theoretical foundation, and miss some opportunities to further accelerate the inference.

\subsection{Concurrent Works}
Recent studies have begun to explore the intersection of Speculative Decoding and MoE. \citet{saxena2025utility} selectively enables speculation to avoid slowdowns and dynamically tunes number of draft tokens to accelerate MoE serving, which overlooks the activation utility generated in a single step. \citet{huang2025moesd} analyzed the efficacy of SD for sparse MoE, yet their theoretical analysis relies on the assumption of uniformly activated experts, ignoring certain correlations among drafted tokens. 
System-oriented works like those by \citet{wang2025accelerating} and \citet{zhuge2025specoffload} utilize CPU resources primarily to interleave the execution of draft and target models to maximize throughput, rather than offloading specific experts to reduce latency.
While \citet{chen2025sp} and \citet{wang2025moe} leverage information from the draft model for activation prediction, they overlook the continuous activation trends across iterations. \citet{bang2026specmoe} remains constrained to GPU-centric execution, failing to fully unlock the potential of heterogeneous computing, despite its self-assisted drafting. \citet{mcdanel2026moe} suffers from precision loss for limiting expert budget during verification.

\section{Future Works}
It is important to note that our work is orthogonal to a broad class of efforts aimed at accelerating speculative decoding itself. Techniques such as KV-cache compression and reuse \citep{sun2024triforce}, as well as retraining- or alignment-based methods that improve draft--target agreement \citep{li2024eagle}, can be readily integrated with \ProjectName{} to further reduce both memory footprint and verification overhead, thereby improving end-to-end efficiency. In addition, self-speculation approaches \citep{zhang2023draft}, which eliminate the need for a separate draft model, can significantly reduce memory consumption and free additional GPU capacity for expert caching, effectively increasing the achievable expert cache ratio within our framework. Our workload balancer can further take KV cache offloading and scheduling into account~\citep{qin2024mooncake, jaillet2025online}, to avoid overhead caused by KV cache from both draft model and target model.

Beyond standard MoE architectures, emerging sparse model designs open new opportunities for utility-guided scheduling. Advanced architectures such as Mixture-of-Lookup-Experts (MoLE) \citep{jie2025mixture} and conditional memory systems like Engram \citep{Cheng2026ConditionalMemory} introduce new axes of sparsity and structured lookup mechanisms. These designs naturally expose richer and more stable access patterns, which could enable more accurate expert utility estimation and prefetching decisions. As models become increasingly sparse across parameters, activations, and memory accesses, the speculative-utility-based scheduling paradigm proposed in this work is expected to play an increasingly central role in efficient inference systems.

\section{Limitations}\label{sec:lim}
Despite its effectiveness, our approach has several limitations. First, \ProjectName{} fundamentally relies on speculative decoding to provide informative, frequency-valued activation signals; when speculative decoding is ineffective (e.g., due to low acceptance rates or strict latency constraints), the benefits of speculative utility estimation are reduced. Second, the proposed system introduces additional scheduling and bookkeeping complexity, including utility tracking, online threshold optimization, and asynchronous execution management, which may incur nontrivial engineering complexity in practice.


\section{Broader Impacts}\label{sec:impact}

This work improves the efficiency of Mixture-of-Experts (MoE) inference in resource-constrained settings. 
A potential positive impact is to lower the hardware and energy cost of serving large sparse models, which may broaden access to advanced language models on commodity devices and edge platforms.

At the same time, improved inference efficiency can also facilitate wider deployment of such models in downstream applications, including those where generated content may be misleading, inappropriate, or over-relied upon. 
Our work does not introduce new model capabilities or safety mechanisms; it is a systems optimization method for existing models. 
Therefore, the societal impact of this work depends primarily on how the underlying models are deployed and governed in practice.

We hope this work contributes to more efficient and accessible AI systems, while recognizing that efficiency gains should be accompanied by appropriate safety evaluation and application-level oversight.


\newpage
\section*{NeurIPS Paper Checklist}

\begin{enumerate}

\item {\bf Claims}
    \item[] Question: Do the main claims made in the abstract and introduction accurately reflect the paper's contributions and scope?
    \item[] Answer: \answerYes{}
    \item[] Justification: The abstract and introduction state the main contributions of repurposing speculative decoding as a lookahead signal for heterogeneous expert scheduling, and these claims are supported by both the theoretical analysis and experimental results reported in the paper.

\item {\bf Limitations}
    \item[] Question: Does the paper discuss the limitations of the work performed by the authors?
    \item[] Answer: \answerYes{}
    \item[] Justification: The paper includes a dedicated \textit{Limitations} discussion in Appendix~\ref{sec:lim}, covering reliance on speculative decoding effectiveness, additional scheduling/bookkeeping complexity, and the current focus on single-GPU, batch-size-one edge scenarios.

\item {\bf Theory assumptions and proofs}
    \item[] Question: For each theoretical result, does the paper provide the full set of assumptions and a complete (and correct) proof?
    \item[] Answer: \answerYes{}
    \item[] Justification: The paper provides formal assumptions and derivations for the theoretical analysis in the appendix, including the analysis of expert reuse, information gain, fault tolerance, and the bounded-drift justification for utility estimation.

\item {\bf Experimental result reproducibility}
    \item[] Question: Does the paper fully disclose all the information needed to reproduce the main experimental results of the paper to the extent that it affects the main claims and/or conclusions of the paper (regardless of whether the code and data are provided or not)?
    \item[] Answer: \answerYes{}
    \item[] Justification: The paper specifies the evaluated model setup, hardware platform, benchmark suite, key hyperparameters, baselines, and evaluation metrics, and code is publicly available, providing a reasonable path to reproduce the main empirical results.

\item {\bf Open access to data and code}
    \item[] Question: Does the paper provide open access to the data and code, with sufficient instructions to faithfully reproduce the main experimental results, as described in supplemental material?
    \item[] Answer: \answerYes{} 
    \item[] Justification: Code is available at \url{https://github.com/lshAlgorithm/MoE-SpAc}. The experiments are conducted on public benchmarks and existing open or accessible model assets.
    
\item {\bf Experimental setting/details}
    \item[] Question: Does the paper specify all the training and test details (e.g., data splits, hyperparameters, how they were chosen, type of optimizer) necessary to understand the results?
    \item[] Answer: \answerYes{}
    \item[] Justification: The paper provides the experimental hardware, benchmark suite, baselines, target and draft models, batch size, draft length, utility upper bound, forgetting factor, and evaluation metrics needed to understand the reported inference results. There is no model training in the proposed method itself.

\item {\bf Experiment statistical significance}
    \item[] Question: Does the paper report error bars suitably and correctly defined or other appropriate information about the statistical significance of the experiments?
    \item[] Answer: \answerYes{}
    \item[] Justification: We provide the results with five repeated experiments in tables. In Figure~\ref{fig:online} and Figure~\ref{fig:batch_size}, we report the error bar in colors with opacity.

\item {\bf Experiments compute resources}
\item[] Question: For each experiment, does the paper provide sufficient information on the computer resources (type of compute workers, memory, time of execution) needed to reproduce the experiments?
\item[] Answer: \answerYes{} 
\item[] Justification: The paper specifies the main hardware platform, including a single NVIDIA GeForce RTX 4090 GPU and PCIe 4.0 CPU--GPU interconnect. Memory and time of execution are all reported in Section~\ref{sec:exp}.

\item {\bf Code of ethics}
\item[] Question: Does the research conducted in the paper conform, in every respect, with the NeurIPS Code of Ethics \url{https://neurips.cc/public/EthicsGuidelines}?
\item[] Answer: \answerYes{}
\item[] Justification: The work is a systems and inference-efficiency study conducted on existing models and benchmarks, and the authors are not aware of any aspect of the research that conflicts with the NeurIPS Code of Ethics.

\item {\bf Broader impacts}
\item[] Question: Does the paper discuss both potential positive societal impacts and negative societal impacts of the work performed?
\item[] Answer: \answerYes{} 
\item[] Justification: Appendix~\ref{sec:impact} states that potential positive impacts include making large MoE models more accessible on resource-constrained devices, while possible negative impacts include enabling more efficient deployment of powerful language models in misuse scenarios.

\item {\bf Safeguards}
\item[] Question: Does the paper describe safeguards that have been put in place for responsible release of data or models that have a high risk for misuse (e.g., pre-trained language models, image generators, or scraped datasets)?
\item[] Answer: \answerNA{}
\item[] Justification: The paper does not release a new generative model or dataset; it proposes an inference framework built on existing models and benchmarks. Any model-specific safeguards are inherited from the original model providers.

\item {\bf Licenses for existing assets}
\item[] Question: Are the creators or original owners of assets (e.g., code, data, models), used in the paper, properly credited and are the license and terms of use explicitly mentioned and properly respected?
\item[] Answer: \answerYes{} 
\item[] Justification: The paper cites the existing benchmarks, models, and baselines used in the experiments. 

\item {\bf New assets}
\item[] Question: Are new assets introduced in the paper well documented and is the documentation provided alongside the assets?
\item[] Answer: \answerYes{} 
\item[] Justification: The main new asset is the implementation of the proposed inference framework. The paper points to the public code repository.

\item {\bf Crowdsourcing and research with human subjects}
\item[] Question: For crowdsourcing experiments and research with human subjects, does the paper include the full text of instructions given to participants and screenshots, if applicable, as well as details about compensation (if any)?
\item[] Answer: \answerNA{}
\item[] Justification: The paper does not involve crowdsourcing or research with human subjects.

\item {\bf Institutional review board (IRB) approvals or equivalent for research with human subjects}
\item[] Question: Does the paper describe potential risks incurred by study participants, whether such risks were disclosed to the subjects, and whether Institutional Review Board (IRB) approvals (or an equivalent approval/review based on the requirements of your country or institution) were obtained?
\item[] Answer: \answerNA{}
\item[] Justification: The paper does not involve human-subject research and therefore does not require IRB approval.

\item {\bf Declaration of LLM usage}
\item[] Question: Does the paper describe the usage of LLMs if it is an important, original, or non-standard component of the core methods in this research? Note that if the LLM is used only for writing, editing, or formatting purposes and does \emph{not} impact the core methodology, scientific rigor, or originality of the research, declaration is not required.
\item[] Answer: \answerYes{}
\item[] Justification: The work studies inference for existing large language models and explicitly uses a target MoE model and a draft model as part of the experimental and methodological setup. These model usages are described in the paper's experiment section and are central to the evaluation setting. LLMs are used solely as the research target in this paper.

\end{enumerate}

\end{document}